\RequirePackage{fix-cm}
\documentclass[smallextended]{svjour3}       
\smartqed  
\usepackage{graphicx}
\usepackage{cite}
\usepackage{amsmath,amssymb,amsfonts}
\usepackage{algorithmic}
\usepackage{graphicx}
\usepackage{textcomp}
\usepackage{xcolor}
\usepackage{subcaption,threeparttable}
\usepackage{comment}
\usepackage[font=small]{caption}
\usepackage[ruled,longend,linesnumbered]{algorithm2e}

\usepackage{tikz}
\usepackage{multirow}
\usepackage{makecell}
\usepackage{bm}
\usepackage{url}
\usepackage{lscape}
\usepackage{natbib}
\usepackage[bottom]{footmisc}
\usetikzlibrary{plotmarks}

\begin{document}
\SetKwInOut{Input}{Input}
\SetKwInOut{Require}{Require}
\SetKwInOut{Output}{Output}
\SetKwInOut{Initialize}{Initialize}
\SetKwComment{Comment}{$\triangleright$\ }{}

\title{Wisdom of the Contexts: Active Ensemble Learning for Contextual Anomaly Detection}


\author{Ece Calikus         \and S\l{}awomir Nowaczyk \and
        Mohamed-Rafik Bouguelia \and
        Onur Dikmen 
}


\institute{Authors are at Center for Applied Intelligent Systems Research (CAISR),\\ Halmstad University, Box 823, 301 18 Halmstad, Sweden\\
	\email{ece.calikus@hh.se; slawomir.nowaczyk@hh.se; mohamed-rafik.bouguelia@hh.se; onur.dikmen@hh.se}
}

\date{Received: date / Accepted: date}

\maketitle

\begin{abstract}
In contextual anomaly detection, an object is only considered anomalous within a specific context. Most existing methods use a single context based on a set of user-specified contextual features. However, identifying the right context can be very challenging in practice, especially in datasets with a large number of attributes. Furthermore, in real-world systems, there might be multiple anomalies that occur in different contexts and, therefore, require a combination of several ``useful'' contexts to unveil them. In this work, we propose a novel approach, called WisCon (Wisdom of the Contexts), to effectively detect complex contextual anomalies in situations where the true contextual and behavioral attributes are unknown. Our method constructs an ensemble of multiple contexts, with varying importance scores, based on the assumption that not all useful contexts are equally so. We estimate the importance of each context using an active learning approach with a novel query strategy. Experiments show that WisCon significantly outperforms existing baselines in different categories (i.e., active learning methods, unsupervised contextual and non-contextual anomaly detectors) on 18 datasets. Furthermore, the results support our initial hypothesis that there is no single perfect context that successfully uncovers all kinds of contextual anomalies, and leveraging the ``wisdom'' of multiple contexts is necessary.
\end{abstract}

\section{Introduction}
\label{sec:introduction}
Anomaly detection is the task of identifying patterns or data points that differ from the norm. To identify anomalies, most anomaly detection techniques treat all the features associated with a data point equally. In many real-world applications, while some attributes in a feature set provide direct information on the normality or abnormality of objects (i.e., behavioral attributes), others give clues on environmental factors affecting the system (i.e., contextual attributes). For instance, high energy usage for space heating in a household can be normal in winter, whereas the identical behavior would be abnormal in the summer. In this case, even though the ``ambient temperature'' is not the attribute that directly indicates the abnormality, it can be used to determine whether the energy consumption behavior is as expected under a particular set of condition. 


The goal of contextual (or conditional) anomaly detection is to find objects that are anomalous within certain contexts, but are disguised as normal globally (i.e., in the complete feature space). Contextual attributes are used for defining contexts, while behavioral attributes help to determine whether an object $x$ significantly deviates from other related objects, i.e., those sharing similar contextual information with $x$. The majority of existing techniques address this problem using pre-specified features—most commonly assumed to be either spatial or temporal ones—to define the context \citep{chandola2009anomaly}.


However, identifying the ``true'' contextual and behavioral attributes in complex systems is very difficult in practice and often requires extensive domain knowledge. A context can be specified in many different ways, especially in datasets with a large number of features. Considering that the actual roles of different attributes are unknown in most of the cases, an effective contextual anomaly detection should be able to make an automated decision on the ``best'' context among many possible combinations of different attributes. 

Anomaly detection is mostly formulated in an unsupervised fashion as ground truth information is often absent in practice, and acquiring labels can be prohibitively expensive. Lack of ground truth makes defining the ``right'' context for the problem more challenging, as we cannot verify whether the specified context actually reveals the contextual anomalies. 
 
On the other hand, real-world systems often generate very diverse types of anomalies, and many of them may occur in different situations for different reasons. In such cases, a context that seems ``right'' for discovering a specific anomalous behavior may be completely irrelevant when detecting other anomalies occurring in the same system. As in the previous example, a high level of heat consumption in summer is probably abnormal, and it can be detected when the heat consumption is used as the behavioral attribute, and the ambient temperature is the context. On the other hand, broken valves cause the heating system to increase the hot water flow in the pipes unnecessarily. The high flow rate by itself does not indicate an anomalous behavior if it is a result of high heat consumption, because a large flow is necessary to carry sufficient heat. However, a system that has a much higher flow rate than others \emph{with similar heat consumption} is most likely faulty. In this case, the flow rate indicates the abnormality, and the heat consumption serves as the context. We cannot specify both the ambient temperature and heat consumption as a single context, as the heat consumption was the behavioral attribute in the previous example. To be able to identify both the anomalies, we need two different contexts. Evidently, there is no single ``true'' context to ``rule'' them all. Our main hypothesis is that anomalies in such complex systems can only be detected with a proper combination of multiple contexts. 



However, leveraging multiple contexts effectively is not a trivial task, at least not without knowing which contexts are useful. Solutions that treat all available contexts equally would incorporate incorrect decisions when many uninformative contexts are present. Therefore, we need an approach that carefully decides on important contexts that reveal different kinds of contextual anomalies in the system while eliminating the impact of the irrelevant ones.


In this work, we focus on two major challenges: (1) effectively incorporating multiple contexts, given that usefulness of a context is unknown, (2) effectively estimating whether a context is useful or not, without ground truth information concerning which attributes are contextual vs behavioral. The paper introduces the ``Wisdom of the Contexts'' (WisCon) approach to address the problem of contextual anomaly detection, in which the ``true'' roles of attributes are unknown a priori. WisCon leverages the synergy of two worlds, active learning and ensembles. Active learning is concerned with quantifying how useful or irrelevant a context is in unveiling contextual anomalies with a low labeling cost. Ensemble learning is used to combine decisions over multiple useful contexts with varying importance, instead of relying on a single pre-specified one.

Various active learning methods have been previously developed from different perspectives to decide which samples in a dataset are more informative than the others. However, the concept of informativeness is inherently subjective and depends on the problem at hand, the dataset, and the machine learning model. Our work is different from all prior work in one key aspect—--the purpose of our active learning is to query instances that help distinguish between useful and irrelevant contexts accurately. To the best of our knowledge, there is no existing sampling strategy designed based on this objective. It is inherently different from the use of active learning in, for example, supervised machine learning, where classification uncertainty has been used successfully. In anomaly detection, one of the classes is intrinsically more interesting than the other, which presents unique challenges. Moreover, the fact that WisCon uses queries to evaluate \textit{contexts} adds an extra layer of complexity. 

To fill this gap, we propose a novel query strategy that aims to select samples enabling the most accurate estimation of ``usefulness'' of different contexts. It is a committee-based approach, in which each committee member is a different base anomaly detector built based on a particular context. Our strategy mainly ensures that anomalous samples that cannot be detected under the majority of the contexts and, therefore, cannot gain the confidence of the committee, are queried sufficiently often. We claim that a context successfully revealing these anomalies has a higher probability of being ``useful.'' Using this approach, we can easily discard many irrelevant contexts from the decision making by using a limited number of queries.



In summary, the contributions in this paper are as follows: 
\begin{itemize}
	\item \textbf{WisCon approach:} We propose a novel approach for contextual anomaly detection, namely Wisdom of the Contexts (WisCon), that automatically creates contexts, where true contextual and behavioral attributes are not known beforehand, and constructs an ensemble of multiple contexts with an active learning model, 
	\item \textbf{Sampling strategy:} We propose a new committee-based query strategy, low confidence anomaly (LCA) sampling, designed to select anomalous samples that cannot be detected under the majority of the contexts. This strategy allows us to actively query for the anomalies that provide more information about which contexts are ``relevant'' or ``irrelevant'' so that importance of contexts can be estimated within a small budget.
	\item \textbf{Context Ensembles:} We design an ensemble over the weighted combination of different contexts, in which the results from different contexts are aggregated using their importance scores estimated with active learning and a pruning strategy that eliminates irrelevant contexts.
	\item \textbf{An empirical study:} We conduct comprehensive experiments including statistical comparisons with baselines in different categories; performance comparisons and budget analysis among the different state of the art query strategies and our novel strategy; and a study on showing individual benefits of two core concepts, i.e., active learning and multi-context ensembles, in this problem. 
	
\end{itemize}


The remainder of this paper is organized as follows. In Section 2, we review related studies. We introduce notations and the problem definition in Section 3. In Section 4, we introduce the WisCon approach and present the overall framework. Then, we describe different query strategies used in our approach and detail our novel query strategy in Section 5. In Section 6, we empirically verify the effectiveness of our approach. Finally, we conclude the paper in Section 7.

\textbf{Reproducibility:} The datasets used in the experiments (see Section \ref{Datasets}) are publicly available and the implementation of WisCon is open-sourced at \url{https://github.com/caisr-hh}.

\section{Related Work}
Our study is related to the following four lines of work.

\textbf{Contextual Anomaly Detection:}
Anomaly detection has been extensively studied in the literature, and comprehensive surveys can be found in \citep{aggarwal2015outlier, chandola2009anomaly, zimek2012survey}. Most existing works do not make a distinction between attributes and refer to the whole feature set to indicate anomalous behavior. 

\begin{figure}
\centerline{\includegraphics[width= 0.9
\linewidth]{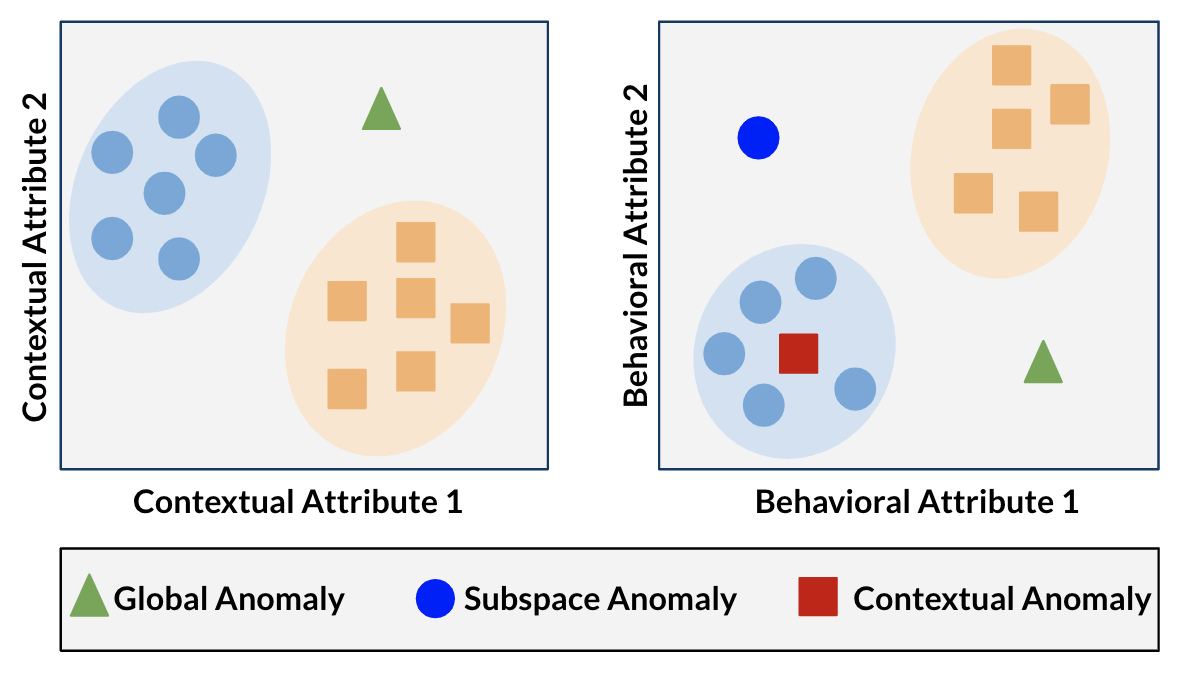}}
\subfloat[\label{con} Contextual space (Context)]{\hspace{.55\linewidth}}
\subfloat[\label{beh} Behavioral space (Behavior)]{\hspace{.4\linewidth}}
\caption{Illustration of different types of anomalies. A global anomaly (blue triangle) stands out in the complete feature space, a subspace anomaly (green circle) is unveiled in a specific subspace, but a contextual anomaly (the red square) can only be detected in behavioral space w.r.t its reference group (other squares) found in contextual space.}
\label{anoms}
\end{figure}

Contextual (or conditional) anomaly detection, which distinguishes contextual (i.e., environmental) attributes from behavioral (i.e., indicator) ones, is relatively new and has not been explored as much as non-contextual methods.  A contextual anomaly is defined as an object that shares similarity on some attributes (i.e., contextual) with a reference group of objects, but deviate dramatically from the reference group on some other attributes (i.e., behavioral) (Fig. \ref{anoms}). The key distinguishing factor is that \textit{exclusively} behavioral attributes are of interest in detecting anomalies -- an object that deviates in terms of contextual attributes should be ignored. 

Early works of contextual anomaly detection have mostly been studied in time-series \citep{weigend1995nonlinear, salvador2004learning} and spatial data \citep{kou2006spatial, shekhar2001detecting}, where either spatial or temporal attributes are considered as context. These methods are usually application-specific and not suitable for other domains where the contexts can vary.       

On the other hand, \cite{song2007conditional} proposed the Conditional Anomaly Detection (CAD) algorithm as a general method to detect contextual (conditional) anomalies. CAD is a probabilistic approach based on the assumption that behavioral attributes conditionally depend on the contextual attributes, and samples that violate this dependence are anomalous. They model contextual and behavioral attributes using Gaussian Mixture Models (GMM). Then, a mapping function among Gaussian components is learned to capture the probabilistic dependence between contextual and behavioral attributes. They have also presented a perturbation schema based on their CAD model, which enables one to artificially inject contextual anomalies into datasets. 

\cite{liang2016robust} proposed Robust Contextual Outlier Detection (ROCOD) addressing the problem of ``sparsity'' in the context space. They argue that it is hard to properly estimate a sample's ``reference group'' (i.e., a group of objects that have similar contextual attribute values) if that sample is highly scattered in the context space. Consequently, its ``outlierness'' in the ``behavioral space'' cannot be determined accurately due to the lack of ``reference''. To tackle this problem, they propose the ROCOD algorithm, which combines the local and global estimation of the behavioral attributes. The local expected behavior of a sample is computed as the average values of behavioral attributes among all samples in its reference group, while the global expected behavior is estimated with a regression model that predicts the values of behavioral attributes by taking the contextual attributes as input. Another method introduced by \cite{zheng2017contextual} accommodates metric learning with the CAD model to effectively measure similarities (or distances) in the contextual feature space. They claim that combining the contextual attributes, including both spatial and non-spatial features, is not a trivial task, and simply using Euclidean distance to estimate reference groups is not an effective solution. Therefore, they proposed to use metric learning to learn the distance metric to find meaningful contextual neighbors. 
However, all of these approaches assume a single, user-defined context and do not accommodate multiple contexts. 

The most closely related prior work is ConOut \citep{meghanath2018conout}, which automatically generates multiple contexts and incorporates them into an iForest \citep{liu2008isolation} based anomaly detection algorithm. The motivation behind this approach is that including all possible contexts is computationally expensive, and contexts that have highly dependent attributes may result in similar reference groups. Therefore, they propose a measure by leveraging statistical tests to quantify dependence between two samples and eliminate redundant contexts from their context set. However, they use a naive aggregation method to combine multiple contexts, where the final anomaly scores are computed by taking the maximum of the anomaly scores across all contexts. They include all generated contexts into the final decision, assuming they are all useful and equally important, which is the core difference between ConOut and our approach.

\textbf{Active Learning:}
Active learning aims to allow learning algorithms to achieve higher performance with few but informative samples, especially when the labels are expensive and difficult to obtain. Active learning has also been applied to anomaly detection problems using different query strategies such as uncertainty sampling \citep{gornitz2013toward,sindhwani_melville_lawrence_2009,calikus2019interactive}, query-by-committee \citep{abe2006outlier}, and querying most likely anomalies \citep{das2016incorporating,siddiqui2018feedback,das2020discovering}.  

\cite{das2016incorporating,das2020discovering} proposed Active Anomaly Discovery (AAD) algorithm, in which the human analyst provides feedback to the algorithm on true labels (nominal or anomaly) and the algorithm updates its parameters to be consistent with the analyst's feedback. At each step, the AAD approach defines and solves an optimization problem based on all prior analyst feedback resulting in new weights for the base detectors. AAD has been implemented with the LODA anomaly detector \citep{pevny2016loda} in \cite{das2016incorporating}, and with iForest in \cite{das2017incorporating}. Unlike other active learning approaches where the goal is to improve the model's overall accuracy with few labeled instances, AAD approach aims to maximize the number of true anomalies presented to the user and adjusts the weights of different ensemble components based on this objective. With a similar goal, \cite{lamba2019learning} presented OJRank, which selects the top instance among the ranked anomalies to be verified by the expert. To reduce the expert's verification effort, they aim to show similar anomalies at the top. Therefore, OJRank re-ranks the anomalies by leveraging the feedback from the user.

\cite{siddiqui2018feedback} proposed a feedback-guided anomaly discovery procedure, in which they also try to keep the most interesting anomalies at the top. In their setting, the analyst provides feedback about whether the top-ranked instance was of interest or not after investigating it. This feedback is then used to adjust the anomaly ranking after every interaction with the analyst, trying to move anomalies of interest closer to the top. None of these approaches above have investigated active learning for contextual anomaly detection. 

Many other active learning methods have been previously developed from different perspectives to decide which samples in a dataset are more informative than the others, such as uncertainty sampling \cite{tong2001support,sindhwani_melville_lawrence_2009}, query-by-committee \cite{seung1992query}, and expected model change \cite{cai2017active}. However, the definition of informativeness depends on the problem and can vary greatly in different applications. For example, several sampling strategies introduced with anomaly detection algorithms explained above focus on maximizing the number of anomalies presented to an analyst \cite{das2016incorporating,lamba2019learning} to reduce the analyst's inspection effort. On the other hand, many other works are interested in samples that lie on the decision boundaries to quickly separate different classes by querying for few labels. 

The key difference between WisCon and these methods is that in our case the goal is to learn the the usefulness of each context; more specifically, given a set of contexts, to assign a specific weight to each and every one. The ``informativeness'' criterion used by previous active learning strategies is not effective for distinguishing between useful and irrelevant contexts, given the fact that valuable contexts are rare. Therefore, we design a new query strategy with a more suitable informativeness metric to assess the importance of different contexts. 



\textbf{Outlier Ensembles:}
Ensemble learning methods are concerned with combining the predictions from different base models to achieve more robust, reliable results. Anomaly detection is an inherently subjective problem, where the performances of detection algorithms vary greatly based on many factors— e.g., the problem domain, characteristics of data, and types of anomalies \citep{calikus2020no}. The performance of a single detector can fluctuate greatly under different applications. Therefore, outlier ensembles have received increasing attention in the research community, focusing on combining multiple models to leverage the ``wisdom of the crowd'' over ``an individual''.  
Most of the previous ensemble approaches introduced in the context of anomaly detection can be categorized into two categories: (i) data-centered and (ii) model-centered. The former combines the different subsets or subspaces of the data using the same algorithm, while the latter combines the anomaly scores from different algorithms built on the same dataset \citep{aggarwal2013outlier}.

As an example of a data-centered outlier ensemble, \cite{liu2008isolation} propose a method inspired by random forests \citep{breiman2001random}, which uses different subsamples of training data to construct an ensemble of trees to isolate outliers. There are most recent techniques based on an ensemble of randomized space trees— RS-Forest \citep{wu2014rs}, rotated bagging with variable subsampling techniques \citep{aggarwal2015theoretical}, projection-based histogram ensemble— LODA \citep{pevny2016loda}, and the randomized subspace hashing ensemble RS-Hash \citep{sathe2016subspace}.

There are also many model-centered ensembles including simpler approaches such as average, maximization, and weighted average \citep{aggarwal2017outlier}, and more complex models like SELECT \citep{rayana2016less}, CARE \citep{rayana2016sequential}, and BoostSelect \citep{campos2018unsupervised}. In addition to unsupervised methods, hybrid or semi-supervised approaches have been proposed to incorporate ground-truth information by \cite{micenkova2015learning}, and \cite{zhao2018xgbod}. 

Unlike the previous work, our study incorporates the idea of effectively combining different contexts with varying importance, rather than combining different anomaly detectors, to handle the diversity of complex contextual anomalies. Learning mixtures of models is more challenging in our case, as useful contexts are rare and unknown. 

\textbf{Subspace Outlier Detection:}
Subspace outlier detection is a closely related problem, as it aims to detect anomalies hidden in high-dimensional feature space. The methods tackling this problem aim at identifying outliers in high-dimensional data using different subspaces of the original space (i.e., a subset of the complete feature set).

One of the earliest works on subspace outlier detection \citep{aggarwal2001outlier} proposed a model in which outliers were defined by axis-parallel subspaces. In this paper, an evolutionary algorithm was proposed to discover the lower-dimensional subspaces in which the outliers may exist. Another method was proposed by \cite{kriegel2009outlier} for distance-based outlier detection with subspace outlier degree. For each object in the dataset, it explores the axis-parallel subspace spanned by its shared neighbors and determines how much the object deviates from these neighbors in this subspace.

There are also ensemble-based methods based on randomly sampling subspaces and combining the scores from different subspaces. They attempt to improve detection robustness by bagging the results from different subsets of features. One of the earliest works for diversifying detectors by random subsets of features is Feature Bagging (FB) \citep{lazarevic2005feature}. It uses LOF algorithm as the underlying base detector. Other ensemble methods such as LODA \citep{pevny2016loda}, rotated bagging \citep{aggarwal2015theoretical}, subspace histograms \citep{sathe2016subspace}, high-contrast subspaces \citep{keller2012hics}, and relevant subspaces \citep{muller2011statistical,keller2013flexible} have been also proposed for subspace outlier detection.

The major difference between subspace and contextual anomaly detection is that subspace methods do not distinguish the attributes as contextual and behavioral. We demonstrate this difference with a toy example in Fig \ref{anoms}. The subspace anomaly (green circle) can be identified just using the second subspace (Fig. \ref{beh}) while being hidden in other subspaces, including the global feature space. On the other hand, the contextual anomaly (red square) cannot be found just by searching for it in different subspaces, as it is masked within a group of points (i.e., circles). Therefore, it requires a contextual detector that effectively identifies its reference group (i.e., other squares) in the contextual space (Fig. \ref{con}) and determines whether it actually deviates from this group in the behavioral space (Fig. \ref{beh}). 
 
\section{Problem Definition}
Suppose we have an unlabeled dataset $U \in \mathbb{R}^{n \times d}$, where $n$ is the number of points and $d$ is the number of features. We denote a point from $U$ as $x \in \mathbb{R}^d$ and the set of real-valued features in $U$ as $F = \{f_1, f_2, . . . , f_d\}$. 




\begin{table}[t]
\caption{Notations used throughout the paper}
\begin{center}
\renewcommand{\arraystretch}{1.2}
\begin{tabular}{l|l}

\textbf{Symbol} & \textbf{Definition}\\
\hline

$U$& Unlabeled dataset\\

$L$& Labeled dataset\\

$x$& Data point, $x \in U$ \\

$(x,y)$& Data point, $(x,y)\in L$ \\

$F$& Set of features, $|F|=d$\\

$C$& Context, $C \subset F$\\
$B$& Behavior, $B \subset F$, $B= F \setminus C$\\

$C^\prime$ & True context\\

$\hat{C}$& Set of contexts extracted from $F$, $C_i \in \hat{C}$\\

$R(C,x)$& Reference group of $x$ w.r.t $C$\\

$s_{i,j}$& Anomaly score of $x_j \in U$ within $C_i\in \hat{C}$, $s_{i,j} \in \lbrack0,1\rbrack$\\

$S_i$& Set of anomaly scores for $C_i\in \hat{C}$, $s_{i,j}\in S_i$\\

$\hat{S}$& Matrix of anomaly scores, $\hat{S} \in \mathbb{R}^{n \times m}$, $n=|U|$, $m=|\hat{C}|$, and $S_i$ is the column of $\hat{S}$ \\

$p_{i,j}$& Prediction of $x_j \in U$ in $C_i\in \hat{C}$, $p_{i,j} \in \{0,1\}$ \\

$O$& Oracle\\

$b$& Budget\\

$\epsilon_i$& Detection error in $C_i$\\

$Q()$ & Query strategy\\

$\theta_j$& Sample weight of $x_j \in U$\\

$I_i$& The importance score of $C_i$\\
\hline
\end{tabular}
\vspace{-1em}
\label{tab1}
\end{center}
\end{table}

\noindent
\begin{definition}[\bfseries Context and Behavior]
\textit{Given a dataset $U \in \mathbb{R}^{n \times d}$, a context $C$ is a set of contextual attributes\footnotemark, where the corresponding set of behavioral attributes is defined as a behavior $B$. We construct a context $C$ such that $C \subset F$, $B = F \setminus C$, $|C| > 0$, $|B| > 0$ and, therefore, $|C| + |B| = d$. }
\end{definition}

\footnotetext{We use the terms ``feature'' and ``attribute'' interchangeably throughout this paper. Since both context and behavior correspond to different attribute spaces, we also alternate the terms ``context'' and ``contextual space'' as well as ``behavior'' and ``behavioral space'' occasionally.}

\begin{definition}[\bfseries True Context]
\textit{True context $C^\prime$ is the context that represents the actual (ground truth) contextual attributes in a dataset.}
\end{definition}


\begin{definition}[\bfseries Reference Group]
\textit{Given a context $C$, and a data point $x \in U$, its reference group $R(C,x) \subset U$ is a group of points in $U$ that share similarity with $x$ w.r.t $C$.} 
\end{definition}

\begin{definition}[\bfseries Contextual Anomaly Score]
\textit{Given a context $C$, a set of behavioral attributes $B$, a data point $x_j \in U$, and its reference group $R(C,x_j)$, $s_j \in \lbrack0,1\rbrack$ is the contextual anomaly score of $x_j$ that quantifies the deviation of $x_j$ from $R(C,x_j)$ w.r.t $B$.}
\end{definition}
\begin{definition}[\bfseries Oracle]
\textit{It is the source of ground truth that provides the true label $y_j \in \{0,1\}$ of a sample $x_j$ queried by the active learning, and is assumed to always return correct labels.} 
\end{definition}



\begin{problem}[\bfseries Active Multi-Context Anomaly Detection]
\textit{\\\textbf{Given:}}
\begin{itemize}
	\item \textit{a dataset of points $U$ described by real valued features $F$, in which the true context $C^\prime$ is unknown apriori,}
	\item \textit{an oracle $O$,}
	\item \textit{and a budget $b$;} 
\end{itemize}
\textit{\textbf{Produce:}}
	\begin{itemize}
	\item \textit{a contextual anomaly score for each point in $U$ after obtaining $b$ labels from $O$, such that the anomaly detection performance is maximized} 
	\end{itemize}
\end{problem}

\section{Wisdom of the Contexts (WisCon) Framework}

\subsection{Framework Overview}
\begin{figure}
\centerline{\includegraphics[width= 1.1\linewidth]{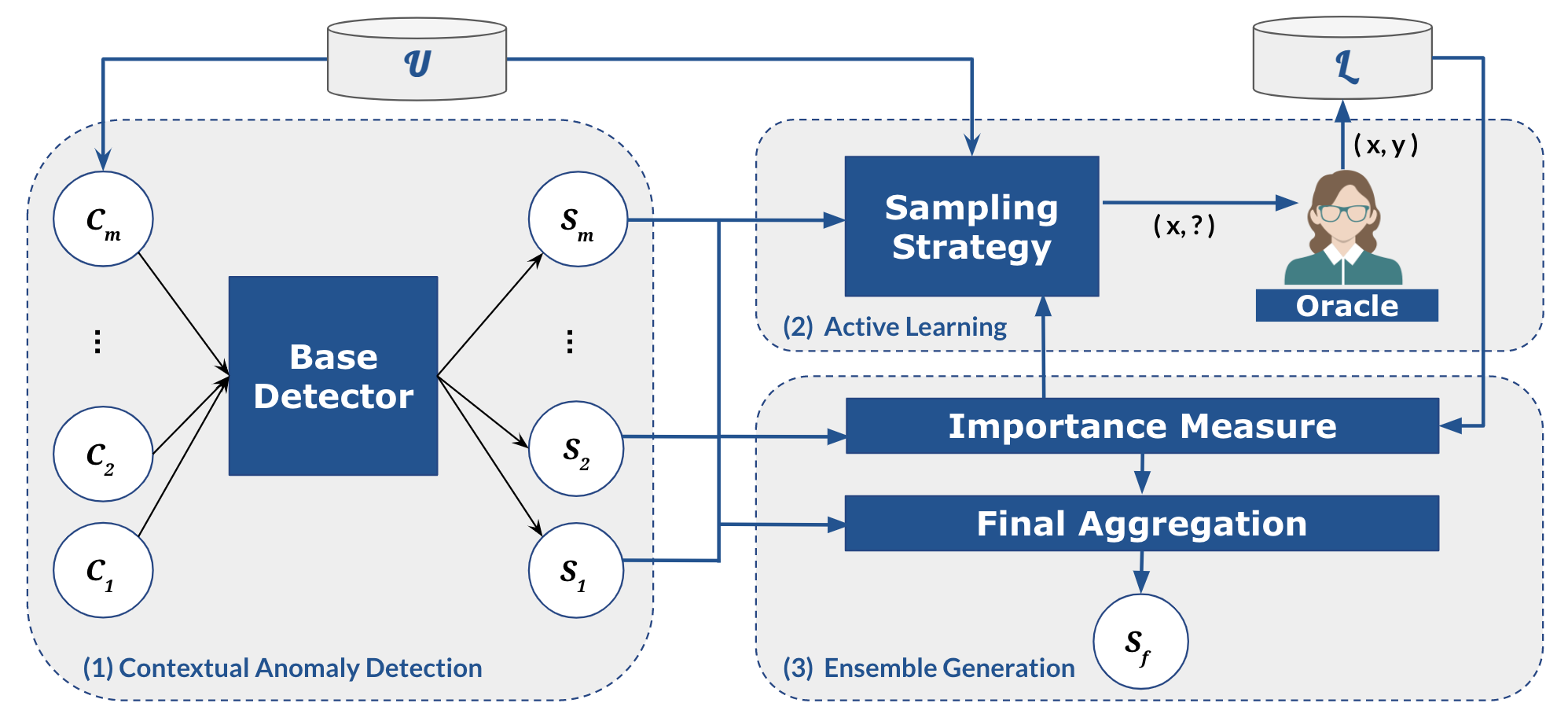}}
\caption{Illustration of the WisCon approach represented with three key building blocks. In the first block, WisCon generates contextual anomaly detectors using different contexts extracted from the feature set (Section \ref{basedetector}). The second building block, active learning, selects the samples to query the oracle for labels, until the budget is exceeded (Section \ref{active_learning}). These labels are iteratively used by the importance measure to estimate importance scores of the contexts (Section \ref{importance_measure}). Finally, WisCon combines the results from different contexts based on their importance scores to produce final anomaly scores (Section \ref{ensemble}).}
\label{framework}
\end{figure}

The proposed WisCon framework consists of three main building blocks. The overall structure is visualized in Fig. \ref{framework}, and the detailed algorithm is presented in Alg. 1. The first block is concerned with the preliminary contextual anomaly detection, in which multiple contexts are created from the feature set, and a base detector for each context produces a set of anomaly scores (Section \ref{basedetector}). The base detector is implemented as a two-step contextual detector, where the first step focuses on discovering the reference groups in data using contextual attributes. The second step measures the deviation from the reference group using the appropriate behavioral attributes. Since the reference group is assumed to include data points similar to each other, any distance-based approach can be used to estimate their similarity in context. 

The second building block is concerned with the active learning schema of the framework (Section \ref{active_learning}). In this work, our goal is to assign different importance to various contexts by rewarding contexts that unveil more anomalies. We leverage active learning to select a small budget of informative instances to be queried for labels such that the importance scores of contexts can be effectively estimated using those labeled instances.

Finally, the third block is ensemble generation and deals with building the ensemble model, which produces final anomaly scores after aggregating the scores from different contexts. This module uses an importance measure (Section \ref{importance_measure}) to estimate the importance scores of each context using labels provided during active learning. Then, it combines the anomaly scores (Section \ref{ensemble}) from individual base detectors that each uses a different context generated in the first building block. The combination approach is based on a pruning strategy, which removes the ``irrelevant contexts'' from the ensemble, and then weighted aggregation of remaining (useful) contexts, in which the weights correspond to their importance scores.  

The details of the methods that are used in this framework are explained in the following subsections.



\begin{algorithm}[htbp]
    \SetAlgoLined
    \Input{Unlabeled data $\mathcal{U}=\{x_1,x_2,...,x_n\}$, \\Context set $\hat{C}=\{C_1,C_2, .., C_i\}$,\\ Budget size $b$}
    \Require{A query strategy $Q$}
    \Output{Final anomaly scores $S=\{s_1,s_2,...,s_n\}$}
    Initialize Labeled data $\mathcal{L}=\emptyset$, a set of initial anomaly scores $\hat{S}=\emptyset$ \\
    \For{$i \leq |\hat{C}|$}{
                     Set $S_i=\emptyset$\\
                     Compute reference groups $R=\{R_1,R_2,...,R_k\}$ by clustering samples in $\mathcal{U}$ using contextual attributes in $C_i$ \textbf{(Sec. \ref{basedetector})}\\
                     \For{$R_t \in R$}{
                     Train an iForest $M$ with samples in $R_t$ using behavioral attributes $B_i=F \setminus C_i$ \textbf{(Sec. \ref{basedetector})}\\
                     Compute anomaly scores for samples in $R_t$ using $M$\\
                     Add anomaly scores to $S_i$}
                     Unify the scores in $S_i$\\
                     Add unified $S_i$ to $\hat{S}$
                }
    Set sample weights $\theta=\emptyset$\\ 
    \While{$|\mathcal{L}| \leq b$}{
     Select $x_j$ by maximizing the objective in \textbf{(Eq. 1)} w.r.t $Q$\\
     Get label $y_j$ \textbf{\{anomaly, normal}\} on $x_j$\\
     Compute the sample weight $\theta_j$ of $(x_j,y_j)$ w.r.t $Q$ \textbf{(Sec. \ref{queries})}\\
     Remove $x_j$ from $\mathcal{U}$\\
     Add $(x_j,y_j)$ to $\mathcal{L}$\\
     Add $\theta_j$ to $\theta$ \\
     \For{$C_i$ in $\hat{C}$}{
        Estimate detection error $\epsilon_i$ for $C_i$ using $S_i$ and $\theta$ \textbf{(Eq. 2)}\\ 
        Update importance score $I_i$ of $C_i$ using $\epsilon_i$ \textbf{(Eq. 3)}
    }
    }
    Prune anomaly scores in $\hat{S}$ using importance scores of $\hat{C}$  \textbf{(Sec. \ref{ensemble})}\\
    Compute final anomaly scores $S$ by importance weighted aggregation of pruned $\hat{S}_{p}$ \textbf{(Eq. 4)} 
    
\Return Final anomaly scores $S$\;
\caption{WisCon Algorithm}
\label{sampling}
\end{algorithm} 

\subsection{Base Detector} \label{basedetector}
%

Assume we are given an unlabeled dataset $U \in \mathbb{R}^{n \times d}$ with features $F=\{f_1,f_2,...,f_d\}$ and a set of contexts $\hat{C}=\{C_1,C_2,...,C_m\}$ that is obtained from $F$. For each context $C_i \in \hat{C}$, we create an individual base detector that assigns an anomaly score to each sample $x_j \in U$ using two steps: (i) estimating reference groups (in the contextual space) and (ii) measuring deviations from these reference groups (in the behavioral space). More specifically, our base detector first clusters the instances using their contextual attributes; each cluster is a reference group. Then, second, it uses an anomaly detection method to find deviations, among data points in each cluster, using behavioral attributes.

Formally, for any $C_i \in \hat{C}$, the base detector estimates the reference groups by clustering samples in $U$ w.r.t the contextual attributes in $C_i$. Suppose we represent the original dataset $U$ as $U=(U^{(C)},U^{(B)})$, where a data point $x = (x^{(C)} ,x^{(B)})$ is composed of a contextual attribute vector $x^{(C)} \in \mathbb{R}^k$ and a behavioral attribute vector $x^{(B)} \in \mathbb{R}^{d-k}$. We cluster the samples only using their contextual attribute vectors in $U^{(C)}$. One can choose any clustering algorithm at this stage, based on the suitability for the underlying structure of their data—for example, high dimensional, overlapping, scattered—to effectively estimate reference groups. Without loss of generality, this paper adopts X-means algorithm \citep{pelleg2000x} because of its simplicity, fast computation, and the automatic decision on the number of clusters. Intuitively, the reference group $R(x_j,C_i)$ of a sample $x_j$ under the context $C_i$, is the set of all samples belonging to the same cluster as $x_j^{(C_i)}$. 

 

After identifying the reference groups, an anomaly detection method is applied to score anomalies based on the deviation from their reference groups in the behavioral space. In this work, we use isolation forest (iForest) algorithm \cite{liu2008isolation} to assign an anomaly score to each point $x_j \in U$. We train separate iForests with the samples in each cluster found in $C_i$ using their behavioral attributes set $B_i = F \setminus C_i$. The results from the iForest give us a set of anomaly scores $S_i$, where $s_{i,j} \in S_i$ is the anomaly score of $x_j$ and $|S_i| = n$.

We also apply a unification step on $S_i$ since base detectors using different contexts can provide anomaly scores with varying range and scale. Unification \citep{kriegel2011interpreting} converts the scores into probability estimates allowing for an effective combination of anomaly scores during the ensemble generation process. We wish to unify the scores for the entire dataset. Therefore, unification is applied to all scores produced by the base detector for the given context, not to individual iForests created for each reference group corresponding to a subset of data. After computing unified anomaly scores $S_i$ for each $C_i \in \hat{C}$, we obtain $\hat{S}$, such that $S_i \in \hat{S}$ and $|\hat{S}|=|\hat{C}|$. 

Any contextual anomaly detector (i.e., a method that takes contextual and behavioral attributes as an input) can be utilized as an individual base detector within our WisCon approach. Alternatively, any ``regular'' anomaly detector can be used after estimating the reference group of any given sample $x_j$. Moreover, other methods for finding the reference group can be used instead of the X-means clustering approach mentioned earlier in this section.

\subsection{Active Learning} \label{active_learning}


WisCon's initial setting is purely unsupervised, since base detectors in the first building block only use an unlabeled dataset $U$ when computing the anomaly scores. In the second step our active learning schema takes these anomaly scores, generated per context, as input and selects the instances for querying the oracle. Our objective is to assign different importance to various contexts by rewarding contexts in which the base detectors make fewer mistakes. The purpose of active learning is to query instances that help distinguish between these useful and irrelevant contexts accurately. 

We utilize the pool-based setting, in which a large pool of unlabeled data $U$ is available at the beginning of the process, and the informativeness of all instances in $U$ is considered when deciding which instances to select.

Until the given budget $b$ is exhausted, the active learning model picks a sample $x_j$ from $U$ with the query strategy $Q$, asks the oracle, receives label $y_j$, and adds $(x_j,y_j)$ to the labeled set $L$. Each time a new label is acquired, WisCon updates the importance scores of the contexts in $\hat{C}$, such that each importance score of $C_i \in \hat{C}$ quantifies how good $C_i$ is in terms of unveiling contextual anomalies (see Section \ref{importance_measure}). Then, WisCon leverages this new knowledge to decide which samples to query next. 



Given unlabeled dataset $U$, a set of anomaly scores $\hat{S}=\{S_1, S_2, ..., S_m\}$ produced by base detectors using contexts $\hat{C}= \{C_1, C_2, ..., C_m\}$, and the budget $b$, the active learning model aims to select the most informative sample to be labeled at each iteration $t\leq b$ by maximizing following objective:
\begin{equation}
\text{maximize}\ E[Q(U,\hat{S})] \text{ subject to } b
\label{eq1}\end{equation}
where $Q(U,\hat{S})$ is a query strategy that returns the sample $x_j$ to be queried to the oracle, and $E[Q(U,\hat{S})]$ is the expected information gain of $x_j$. 

$Q(U,\hat{S})$ is a function that can be replaced with any active selection strategy that uses criteria to measure the ``informativeness'' of selected candidate instances. We implement several state-of-the-art query strategies within WisCon and also propose a novel strategy, i.e., low confidence anomaly sampling. The details of these methods are described in Section \ref{queries}.



\subsection{Importance Measure} \label{importance_measure}
The final anomaly scores are based on the aggregation of information from many different contexts. Therefore, it is crucial to determine which of the possible contexts are highly relevant, and which should be weighted low. We assume irrelevant contexts fail to unveil true anomalies, producing unimportant ones instead; therefore, the base detector using any of these contexts would exhibit high number of false positives. Our idea behind the measure presented below is to reward useful contexts while penalizing irrelevant ones -- based on labels obtained using active learning. It is crucial to note that our active learning oracle is unique as it does not provide direct information about \textit{a context}; instead, it can only be asked about individual data samples, and results of multiple such queries must be aggregated to measure usefulness of any given context.

To determine the importance scores of different contexts within the feature set $F$, we adopt the weighting scheme of AdaBoost \citep{freund1997decision} as our importance measure. It assigns weights for the final ensemble to the base classifiers based on their estimated classification errors. Inspired by this heuristic, we first estimate the ``detection error'' corresponding each individual context using anomaly scores computed by the base detector using that context. To compute the detection error for the given context $C_i$, we convert anomaly scores $S_i=\{s_{i,1},s_{i,2}, .., s_{i,n}\}$ to predictions $P_i=\{p_{i,1},p_{i,2}, .., p_{i,n}\}$ where $p_{i,j} \in \{0,1\}$ using a cutoff between anomalies and nominals. We set the threshold as $0.9$ on the unified anomaly scores to maintain false alarm rate below $10\%$.


Given a labeled dataset $L = \{(x_1,y_1), (x_2,y_2), .., (x_t,y_t)\}$ (from the oracle) and predictions $P_i = \{p_{i,1}, p_{i,2}, .., p_{i,n}\}$, the detection error $\epsilon$ corresponding the context $C_i$ at iteration $t$ can be calculated as:
\begin{equation}
\epsilon_{i,t}= \frac{\sum_{j=1}^{t} \theta_j l_{i,j}}{\sum_{j=1}^{t} \theta_j}
\end{equation}

where $l_{i,j}= \begin{cases}
    0, & p_{i,j} = y_j\\
    1, & p_{i,j} \neq y_j
\end{cases}$, and $\theta_j$ is the sample weight for each $(x_j,y_j) \in L$. Sample weights $\theta$ are estimated depending on the query strategies presented in Section \ref{queries}. If a query strategy does not incur any difference between samples, we simply take $\theta_j=\frac{1}{t}$, for each $(x_j,y_j) \in L$. While $\epsilon_i$ is a direct measure of the objective performance of each context, based on its agreement with the oracle, it is not suitable for direct use as relative context weight. Therefore we take inspiration from AdaBoost.

In the original AdaBoost algorithm, sample weights are determined by giving higher weights to instances that are more difficult to classify. We reuse the overall idea, except we apply it to \textit{context} weights, so that the more successful contexts can have higher relevance. Thus, the importance measure that assigns the importance scores to corresponding contexts can be formulated using the following equation:

\begin{equation}
I_i= \frac{1}{2} ln \left( \frac{1-\epsilon_{i,b}}{\epsilon_{i,b}} \right)
\end{equation}
where $\epsilon_{i,b}$ is the detection error $\epsilon$ for the context $C_i$ at budget $b$.

Contexts assigned with a higher importance score are expected to reveal the contextual anomalies better than other contexts. Furthermore, a negative importance score indicates that the base detector using that context achieves worse performance than a random detector. 

\subsection{Ensemble Pruning and Final Aggregation}\label{ensemble}
As shown in previous works \citep{klementiev2007unsupervised, rayana2014ensemble}, the existence of poor models in the ensemble can be detrimental to the overall performance, or at least contributes to increasing the memory requirements due to storing irrelevant information. Following \cite{rayana2016sequential}, we use a simple pruning strategy, in which we discard a context $C_i$ from the final ensemble if it has a negative importance score, $I_i < 0$. 


We produce the final anomaly score of a sample in $U$ by using weighted aggregation of anomaly scores from the contexts remaining after pruning, where the weights correspond to their importance scores (Section~\ref{importance_measure}). Given the pruned set of anomaly scores $\hat{S}_{p}= \{S_1, S_2, ..., S_P\}$, the final anomaly score $s_j$ of a data point $x_j \in U$ is computed as follows:

\begin{equation}
s_j= \frac{\sum_{i=1}^{P} I_i \times s_{i,j}}{\sum_{i=1}^{P} I_i}
\end{equation}
where $P = |\hat{S}_{p}|$, $I_i$ is the importance score of $C_i$, and $s_{i,j} \in S_i$ is the anomaly score of $x_j$ in $C_i$.

\section{Query Strategies}\label{queries}
In this section, we first present different existing query strategies that are implemented within WisCon (Sections \ref{random}, \ref{qbc}, and \ref{majority_vote}), and then introduce our proposed strategy in Section \ref{lca}. All query strategies described here share the same active learning procedure described in Section \ref{active_learning}: until the given budget $b$ is exhausted, a query strategy $Q$ returns a sample $x \in U$, maximizing the informativeness criterion (Equation~\ref{eq1}) in each iteration, to query the oracle.

\subsection{Random Sampling}\label{random}
 In this strategy, a sample to be queried $x$ is drawn uniformly at random from the unlabeled data $U$ at each iteration $t\leq b$. Since the probability of being sampled for each $x_j \in U$ is equal, it assumes every sample has equal importance, unlike active sampling strategies. Our goal is to improve over this baseline with a more effective selection of ``informative'' samples to be labeled by the oracle.   

\subsection{Query by Committee}\label{qbc}
Query-by-Committee (QBC) is a popular method for active sampling in which a committee of conflicting models is maintained, and samples that cause maximum disagreement among the committee members are selected as the most informative. The idea behind this approach is that these samples are guaranteed to provide high impact per query, regardless of what their true labels turn out to be.

The QBC approaches involve two main steps: (i) constructing the committee of models and (ii) a disagreement measure. In this work, we build the committee with the contextual anomaly detectors that are generated using different contexts and use anomaly scores that are produced by these detectors as class probabilities for the samples. Those probabilities are used along with a measure of disagreement to select the most informative instance to query. Here, we use two traditional disagreement measures with this strategy: (i) consensus entropy and (ii) Kullback-Leibler (KL) Divergence. 

The consensus entropy (also known as soft vote entropy) was first introduced by \cite{dagan1995committee}.
Suppose that $\hat{C} = \{C_1,C_2,...,C_m\}$ is an ensemble with $m$ contexts, $x_j \in U$ is an unlabeled sample with an anomaly score $s_{i,j}\in S_i$ produced in $C_i$ and $Pr_i(y|x)= \begin{cases}
    s_{i,j}, & \text{if } y = 1\\
    1- s_{i,j}, & \text{otherwise} \end{cases}$ is the class probability. Then, the consensus entropy measure is defined as
\begin{equation}
Q_{CE}= \arg\max_{x} - \sum_{y} Pr_C(y|x)log Pr_C(y|x),
\end{equation}
where $Pr_C(y|x)= \dfrac{\sum_{i=1}^{m} I_i \times Pr_i(y|x)}{\sum_{i=1}^{m} I_i}$ is the average, or ``consensus,'' probability that $y$ is the correct label according to the committee. Thus, the query strategy returns the sample that produces the maximum entropy for given ``consensus'' probabilities. This measure can be seen as ensemble generalization of entropy-based uncertainty sampling \citep{settles2012active}.

The second disagreement measure is based on the average Kullback-Leibler (KL) Divergence, which is an information-theoretic approach to calculate the difference between two probability distributions \citep{mccallumzy1998employing}. The disagreement is quantified as the average divergence of each committee member's prediction probabilities from the consensus probability.
\begin{equation}
Q_{KL}= \arg\max_{x} \sum_{i=1}^{m} KL(Pr_i(y|x)||Pr_C(y|x))
\end{equation}
where KL divergence is defined as: \[KL(Pr_i(y|x)||Pr_C(y|x))= \sum_{y} Pr_i(y|x)log \dfrac{Pr_i(y|x)}{Pr_C(y|x)}.\]

In this case, the query strategy considers the sample with the maximum average difference between the label distributions of any committee member and the consensus as the most informative.

\subsection{Most-likely Anomalous Sampling}\label{majority_vote}
In this approach, the major goal is to maximize the total number of anomalous samples verified by the expert within the budget $b$ \cite{das2019active}. Unlike traditional disagreement-based QBC approaches, this strategy targets samples for which the committee has the highest confidence to be true anomalies. Following this goal, we use the simple majority voting rule to pick the sample revealed as an anomaly across the maximum number of contexts. 

Given an ensemble $\hat{C} = \{C_1,C_2,...,C_m\}$ with $m$ contexts, the most-likely anomalous (MLA) sampling is defined as
\begin{equation} 
Q_{MLA}= \arg\max_{x} \dfrac{\sum_{i=1}^{m} I_i \times p_i}{\sum_{i=1}^{m} I_i}
\end{equation}
where  $p_i= \begin{cases}
    1, & \text{if } s_i \geq \text{th}\\
    0, & \text{otherwise} \end{cases}$ is the prediction of $x$ under the context $C_i$. The threshold $th$ is set to $0.9$ throughout this paper as explained in Section \ref{importance_measure}.



\subsection{Low Confidence Anomaly Sampling}\label{lca}
Finally, we propose a new query strategy, low confidence anomaly sampling, which aims at selecting samples that maximize the information gain on the ``usefulness'' of different contexts. It is also an example of a committee-based approach, in which the committee members are base detectors that make decisions under different contexts. The rationale behind this approach, which differentiates it from existing strategies, is that there are multiple useful contexts that actually reveal different anomalies in a system; however, these useful ones are still rare among all available contexts extracted from the feature set. This means that the disagreement, which is the key property of QBC, is inappropriate in our setting; instead, we assume majority of the committee members are in agreement, as only few of them are capable of discovering any given anomaly.

\begin{figure}%
\centering
\begin{subfigure}{0.47\columnwidth}
\includegraphics[width=\columnwidth]{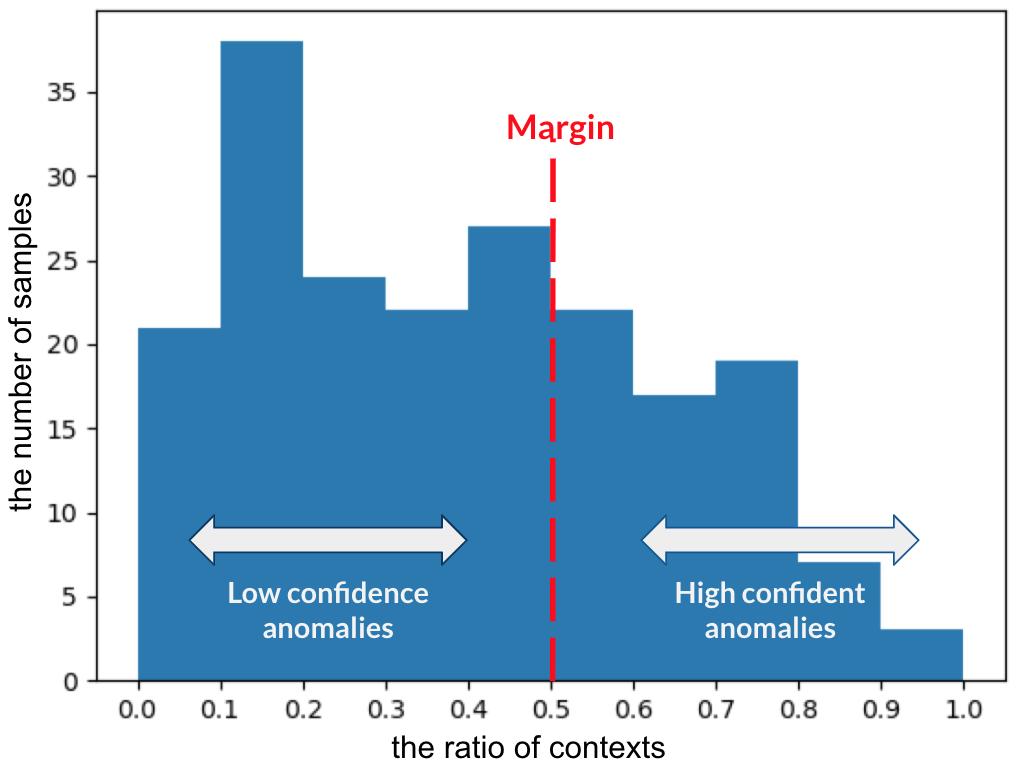}%
\caption{Anomalous samples}%
\label{hist1}%
\end{subfigure}\hfill%
\begin{subfigure}{0.47\columnwidth}
\includegraphics[width=\columnwidth]{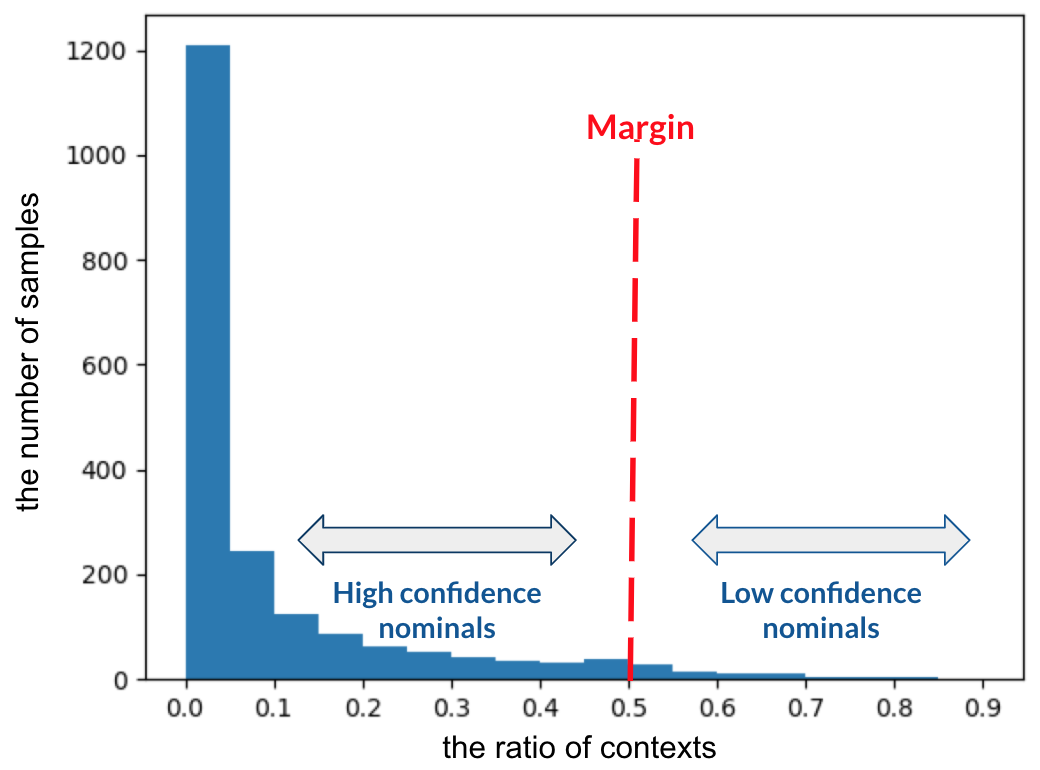}%
\caption{Normal samples}%
\label{hist2}%
\end{subfigure}\hfill%
\caption{The distributions of the ground truth anomalous (a) and normal (b) samples in the Synthetic dataset that are detected as an anomaly by base detectors. Each bin in a histogram covers a range of values corresponding to the ratios of contexts, in which the samples are identified as anomaly, and the height of a bar shows the number of samples within a bin. The margin represents samples that are detected as an anomaly in half of the contexts. The samples fall left-side of the distributions in (a) are the low confidence anomalies meaning that they are unveiled by the minority of contexts. On the contrary, in (b), the samples on the left side of the margin are the high confidence nominals, as they are not incorrectly identified in the majority of the contexts.
}
\label{dists}
\end{figure}

To justify this reasoning, Fig.~\ref{hist1} shows the distributions of correctly detected anomalies in the Synthetic dataset containing artificially generated contextual anomalies (see Section \ref{Datasets}) w.r.t. the ratios of the contexts in which they are detected. More specifically, the first bar in Fig.~\ref{hist1} displays that approximately 20 ground-truth anomalies in Synthetic dataset are correctly identified as anomaly in 0\%-10\% of the contexts. Similarly, Fig.~\ref{hist2}, shows the number of ground-truth normal samples that are incorrectly detected as anomaly. In both figures, the margin represents the threshold for the samples that are detected as anomaly in half of all contexts.


According to Fig. \ref{hist1}, the majority of the anomalous samples in the example fall left-side of the margin, meaning that they cannot be detected in more than half of the contexts. There is a considerable number of anomalous samples that are only detected in less than 20\% of the contexts, supporting our intuition about the infrequency of useful contexts. We refer to these samples as low confidence anomalies— because only the minority of the contexts successfully unveil them— and they are the ``samples of interest'' in this strategy. A context belonging to such a minority group is most likely a ``good'' context and should be assigned a high importance score. 

Considering our main targets are the low confidence anomalies, the first idea would be to sample the instances from the left tail, which accommodates the samples predicted as anomalies under the fewest contexts. However, the confident normal samples also lie on the same side (Fig. \ref{hist2}), and we do not have the true labels of the samples before querying to the oracle. Considering that the anomalies are also rare in the datasets, this type of strategy would result in mostly sampling normal instances rather than the informative, low confidence anomalies. 

Therefore, we intend to select samples around the ``margin'', and exploit the importance scores to iteratively move the low confidence anomalies closer to the ``margin,'' increasing their chances to be selected. To formalize this idea, we first define the \textit{margin rate}, which quantifies the ``closeness'' of a sample to the ``margin'' as follows.

Let $x_j \in U$ be an unlabeled sample and $\hat{C} = \{C_1,C_2,...,C_m\}$ be an ensemble with $m$ contexts. The margin rate of $x_j$ is 

\begin{equation}
    margin(x_j) =  \Bigg(1 - \Bigg|\frac{2\sum_{i=1}^{m} I_i\times p_{i,j}}{\sum_{i=1}^{m} I_i} - 1\Bigg|\Bigg)
\end{equation}
where $I_i$ is the importance score of context $C_i$ and $p_{i,j}= \begin{cases}
    1, & \text{if } s_{i,j} \geq 0.9\\
    0, & \text{otherwise} \end{cases}$ is the prediction of $x_j$ under the context $C_i$, such that $margin(x_j) \in \lbrack0,100\rbrack$. 

Instead of returning $b$ samples with maximum margin rates, we also want to pick samples with lower margin rates from time to time, but less frequently, to allow our model to explore.  

Therefore, we define the low confidence anomaly sampling (LCA) measure as:
\begin{equation}
    Q_{LCA} = \arg\max_{x} \frac{exp (\lambda \times margin(x))}{u_x}
\label{eq1}
\end{equation}

where $\lambda$ is the bias factor, and $u_x$ is a random variable uniformly distributed in $\lbrack0,1\rbrack$ drawn independently for each $x$. The bias factor controls how heavily biased the sampling is towards margin rates, such that $\lambda=0$ gives us random sampling. 

With the above formulation, our sampling strategy gives the samples with higher margin rates higher probabilities to be selected. After each iteration, the margin rates are updated with the new importance scores with the goal of increasing the margin rates of low confidence anomalies, to be selected by the query strategy. To ensure keeping the confident anomalies and normal samples still far from the margin while moving the high confidence anomalies closer, we also accommodate a sample weighting schema. Below we demonstrate the idea with a toy example. 

\begin{figure*}[t]
\centerline{\includegraphics[width=\linewidth]{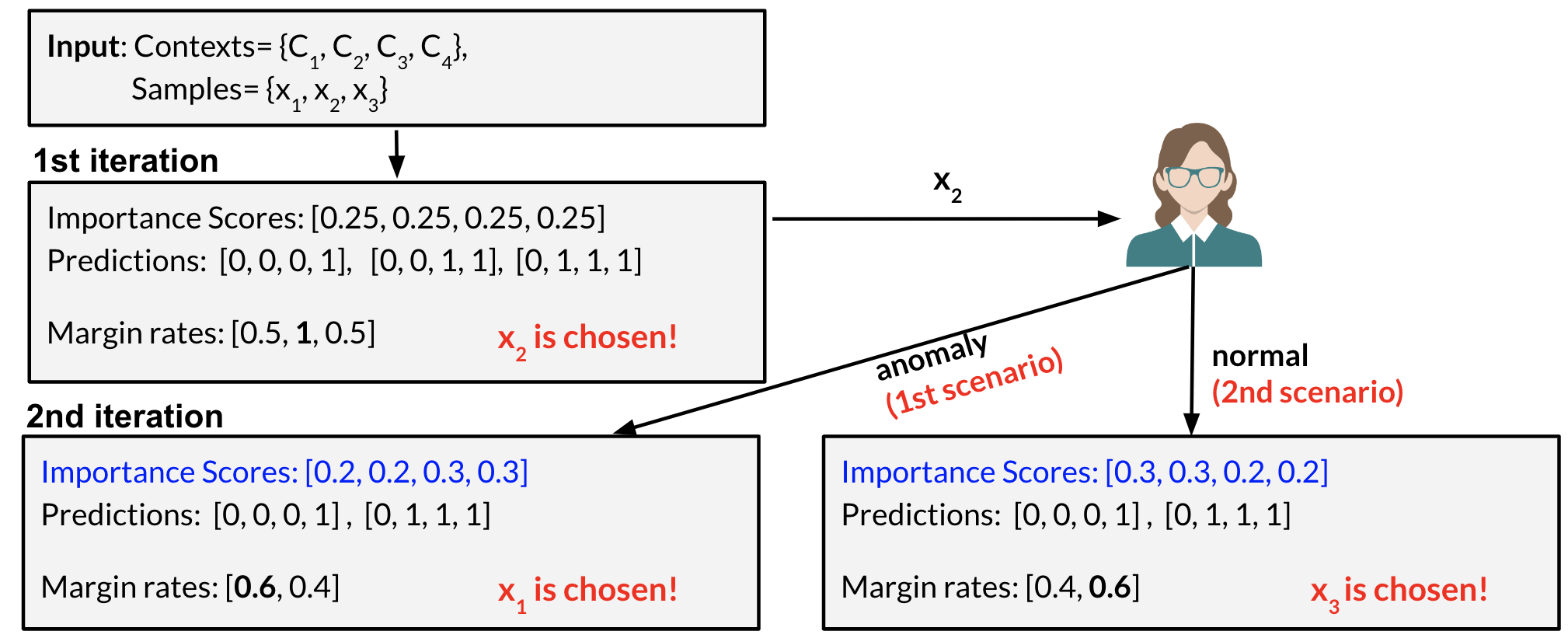}}
\caption{The toy example motivating the need for sample weighting. Given contexts $\{C_1,C_2,C_3,C_4\}$, samples $\{x_1,x_2,x_3\}$ and budget $b=2$, the figure demonstrates the estimation of importance scores according to the samples chosen by LCA sampling under two different scenarios. In the first scenario, $x_2$, the sample chosen after the first iteration is an anomaly, while it is a normal sample in the second scenario. In both scenarios, $x_1$ and $x_3$ are low confidence and high confidence anomalies, respectively.} 
\label{toy_example}
\vspace{-1em}
\end{figure*}

Assume, as shown in Fig. \ref{toy_example}, that we initially have four contexts $\{C_1,C_2,C_3,C_4\}$ with equal importance scores, three samples $\{x_1,x_2,x_3\}$, and budget $b=2$, such that $x_1$ is a low confidence anomaly (detected in only one context) and $x_3$ is a high confidence anomaly (detected in three contexts). There are two scenarios based on the true label of $x_2$.

After the first iteration, the strategy selects the second sample $x_2$, which has the highest margin rate among all the samples. If $x_2$ is an anomaly (first scenario), the importance measure assigns higher importance scores to the third and the fourth contexts since $x_2$ is correctly predicted in these contexts. This selection increases the margin rate of $x_1$ from $0.5$ to $0.6$, while reducing $x_3$'s rate, leading to the selection of $x_1$, the low confidence anomaly, in the second iteration. However, if $x_2$ is a normal sample (second scenario), then the importance measure reduces the weights of $C_3$ and $C_4$, resulting in the opposite scenario where $x_3$, the high confidence anomaly, is selected instead. It should be noted that despite being unlikely, the sample with the lower margin can get selected instead, according to Eq. \ref{eq1}. However, this toy example assumes the most likely outcome from the LCA sampling.    

It can be seen that the second scenario resulted in giving a higher importance score to the context $C_1$, even though it fails to unveil any anomalous samples, over context $C_4$, in which both $x_1$ and $x_3$ are successfully identified. We do not have control over which scenario occurs, as we cannot influence the true labels of the samples picked. However, we can control the \emph{impact} of different samples on context importance by assigning different weights to them after receiving their true labels from the oracle.


Given a sample $(x_j,y_j) \in L$, the sample weight $\theta$ is calculated as:
\begin{equation}
   \theta_j= \begin{cases}
    margin(x_j), & \text{if } y_j = 1\\
    0, & \text{otherwise}
\end{cases}
\end{equation}

As motivated above, normal samples are assigned with $0$ weights so that their impacts are eliminated from the importance scores of the contexts. On the other hand, we give anomalous samples their margin rates as their sample weights. Therefore, the anomalies with higher margin rates also have a stronger influence on the estimation of importance scores in addition to having a higher probability of being selected by the query strategy. By doing that, we sustain a more secure exploration procedure within the LCA sampling, in which the less informative samples have less impact on the importance measure even though they get to be sampled. How the importance measure incorporates the sample weights is described in Section \ref{importance_measure}.

\section{Experiments} \label{experiments}
In this section, we present the experiments carried out to validate WisCon. We start by presenting the datasets used for the experiments (Section \ref{Datasets}), briefly introduce the baselines (Section \ref{Baselines}), and then describe the overall setup (Section \ref{Setup}). Afterward, we present three different experiments conducted to answer the following questions:
\begin{itemize}
    \item \textbf{Q1}. How does the performance of WisCon compare to the performances of the state-of-the-art competitors? (Section \ref{Q1}) 
    \item \textbf{Q2}. How do different query strategies affect the performance of WisCon under different budgets? (Section \ref{Q2}) 
    \item \textbf{Q3}. How beneficial is employing active learning or context ensembles under the WisCon approach for detecting contextual anomalies? (Section \ref{Q3}). 
\end{itemize}

\begin{table*}[t]
\caption{Summary of datasets used in this study}
\renewcommand{\arraystretch}{1.2}
\begin{threeparttable}  
\begin{tabular}{c|cccc}
\hline

\hline
\textbf{Dataset} & \textbf{\textit{Anomalies}}& \textbf{\textit{\#Pts $n$}}& \textbf{\textit{\% Anomalies}}& \textbf{\textit{Dim. $d$}}\\
\hline

\hline
Synthetic 1& Injected& 25250& 250 (1\%) & 10\\

Synthetic 2& Injected& 5100& 100 (2\%) & 10\\

Synthetic 3& Injected& 5100& 50 (2\%) & 50\\

Synthetic 4& Injected& 5100& 50 (2\%) & 10\\

El Nino & Injected& 94784& 1879 (2\%)& 11\\

Houses & Injected& 20640& 413 (2\%)& 9\\

Abalone& Real& 1920& 29 (1.5\%)& 9\\

ANN-Thyroid & Real& 7200& 534 (7.4\%)& 6\\

Arrhythmia & Real & 452 & 66 (15\%) & 274 \\
 
Letter Recognition & Real & 1600 & 100 (6.25\%) & 32\\
Mammography & Real& 11183& 260 (2.32\%)& 6\\
Optdigits & Real & 5216 & 150 (3\%)& 64\\
Pendigits & Real & 6870 & 156 (2.27\%) & 16\\
Satelite & Real & 6435 &2036 (32\%)& 36\\
Satimage & Real & 5803 &71 (1.2\%) &36 \\
Thyroid & Real &  3772 &93 (2.5\%) & 6\\
Vowels & Real& 1456& 50 (3.4\%)& 12\\

Yeast & Real & 1364 & 64 (4.7\%) & 8\\
\hline
\end{tabular}
\end{threeparttable}
\label{dataset_tab}
\end{table*}

\begin{table*}[h!]
\caption{Hyperparameters of baselines}
\renewcommand{\arraystretch}{1.2}
\begin{threeparttable}  
\begin{tabular}{c|cc}
\hline

\hline
\textbf{Baseline} & \textbf{\textit{Parameter}}& \textbf{\textit{Value Set}}\\
\hline

\hline
\multirow{4}{*}{WisCon} & max. no. clusters (XMeans) & \{5, 10, 20\} \\
& no. estimators (iForest)&  100\\
& max. samples (iForest) & 256\\
& $\lambda$ (LCA Sampling) & 0.96\\
\hline
\multirow{4}{*}{AAD} & $\tau$ & 0.03 \\
& $C_A$  & 100\\
& $C_{\xi}$  & 0.001\\
& no. estimators (iForest)&  100\\
& max. samples (iForest) & 256\\

\hline
\multirow{4}{*}{Active-RF} & no. estimators&  \{10, 50, 100, 150, 200\}\\
& max. features & \{auto, sqrt, log2\}\\
& class weight & \{none, balanced, balanced subsample\}\\
& bootstrap & \{true, false\}\\
\hline
Active-KNN & no. neighbors (k) &  \{2, 4, .., 10\} \\
\hline
\multirow{3}{*}{Active-SVM} & $\gamma$ & $10^x$, $x \in \{-4, -3, ..., 4\}$\\
& $C$ &  $10^x$, $x \in \{-2, ..., 4\}$\\
& kernel & \{rbf, polynomial, sigmoid\}\\
\hline
\multirow{1}{*}{ConOut} & $\gamma$ &  \{0.0001,0.001,0.01,0.1,1,10,100,1000\}\\
\hline
\multirow{3}{*}{ROCOD} & no. neighbors (k) &  \{10, 20, .., 100\}\\
& max. depth & \{5, 10, 15, 20\}\\
& min. samples split & \{10, 20, ..., 100\}\\
\hline
CAD & no. components & \{5, 10, 20, 40\} \\
\hline
\multirow{3}{*}{iForest-Con} & max no. clusters &  \{5, 10, 20\} \\
& no. estimators & \{100, 200, ..., 500\}\\
& max features & \{0.1, 0.2, ..., 1\}\\
\hline
\multirow{ 2}{*}{LOF-Con} & max no. clusters &  \{5, 10, 20\}\\
& no. neighbors (k) &  \{10, 20, .., 200\}\\
\hline
\multirow{ 4}{*}{OCSVM-Con} & max no. clusters &  \{5, 10, 20\}\\
& $\gamma$ & $10^x$, $x \in \{-4, -3, ..., 4\}$\\
& $\nu$ &  $\{0.01, 0.5, 0.99\}$\\
& kernel & \{rbf, polynomial, sigmoid\}\\
\hline
\multirow{2}{*}{iForest} & no. estimators & \{100, 200, ..., 500\}\\
& max features & \{0.1, 0.2, ..., 1\}\\
\hline
LOF & no. neighbors (k)  &  \{10, 20, .., 200\}\\
\hline
\multirow{3}{*}{OCSVM}& $\gamma$ & $10^x$, $x \in \{-4, -3, ..., 4\}$\\
& $\nu$ &  $\{0.01, 0.5, 0.99\}$\\
& kernel & \{rbf, polynomial, sigmoid\}\\
\hline
\multirow{2}{*}{LODA} & no. bins & \{2, 4, ..., 100\} \\
& no. random cuts & \{40, 60, ..., 500\}\\
\hline
\multirow{3}{*}{SOD} & no. neighbors (k)  &  \{10, 20, .., 200\}\\
& no. reference set & \{10, 20, .., 100\}\\
& $\alpha$ & \{0.1, 0.2, ..., 1\} \\
\hline
\multirow{2}{*}{Feature Bagging} & no. estimators & \{100, 200, ..., 500\}\\
& max features & \{0.1, 0.2, ..., 1\}\\
\hline
\end{tabular}
\end{threeparttable}
\label{hyperparameters}
\end{table*}

\subsection{Datasets} \label{Datasets}
In this work, we use 18 datasets for evaluation, 14 of which are real datasets, and 12 of them contain labeled ground-truth anomalies of unknown types. For the other two real-datasets without ground-truth anomalies (i.e., El Nino, and Houses), we inject contextual anomalies using the perturbation scheme following \cite{song2007conditional} — a de-facto standard for evaluating contextual anomaly detection methods. The descriptions of the datasets used in this work are provided in Table \ref{dataset_tab}, and the details of the perturbed datasets are described below.

El Nino dataset contains oceanographic and surface meteorological readings taken from a series of buoys positioned throughout the equatorial Pacific. We used the temporal attributes and spatial attributes as contextual attributes (6 attributes) while considering the ones related to winds, humidity, and temperature as behavioral attributes (5 attributes). The dataset has 93,935 points in total, and we further inject 2\% contextual anomalies.

The houses dataset includes information on house prices in California. We used the house price as the behavioral attribute and the other attributes as contextual attributes, such as median income, median housing age, median house value, and so on. That dataset contains 20,640 entries, and we also injected 2\% contextual anomalies.

In addition to real datasets, we use four synthetic datasets with different characteristics. First three Synthetic datasets (Synthetic 1, 2, and 3) include contextual anomalies, while the fourth one (Synthetic 4) has non-contextual, global anomalies. All three contextual datasets consist of the mixtures of multivariate Gaussians using the CAD model \citep{song2007conditional}. The centroids of the Gaussians are randomly generated and the diagonal of the covariance matrix is set to 1/4 of the average distance between the centroids in each dimension.

The first synthetic dataset (i.e., Synthetic 1) is generated only considering one context in which the number of contextual and behavioral attributes is chosen to be 5, (i.e., $d=10$). The number of Gaussian components for both contextual and behavioral attributes is also set to be 5. In total, we sampled $25,000$ data points and injected 1\% contextual anomalies using the perturbation scheme. 

The second dataset, i.e., Synthetic 2 contains contextual anomalies from multiple contexts. The number of contextual/behavioral attributes for three different contexts are set to be $5/5$, $6/4$ and $7/3$, where $d=10$. The number of Gaussian components for all three contexts and corresponding behavioral attributes are set to be $5$ as well. We generated $5,000$ data points for this dataset. However, we injected $500$, $300$ and $200$ contextual anomalies for the first, the second and the third context, respectively. It is important to note that different contexts, therefore, have different importance in this dataset, as each one reveals a different number of anomalies.

The third dataset Synthetic 3 is the higher-dimensional version of Synthetic 1. It consists of 50 features where we set the half of them as contextual and the other half as behavioral attributes. This dataset include $5,100$ data points in total where 2\% of them are injected anomalies. 

Different from the previous synthetic datasets, Synthetic 4 only contains non-contextual, global anomalies. This dataset is included in the experiments to show how WisCon performs when anomalies are not contextual. We use the utility function provided by PyOD\footnote{\url{https://pyod.readthedocs.io/en/latest/}}\citep{zhao2019pyod} library to create this dataset. It includes clusters with varying densities where outliers have significantly lower densities. All clusters are generated as isotropic Gaussian blobs with different sizes. The number of clusters are chosen to be 5, while the number of features is set to 10. This dataset include $5,100$ data points in total where 2\% of them are anomalies.   



The true context for perturbed datasets—Synthetic 1, Synthetic 2, Synthetic 3, El Nino, and Houses, is defined by the set of contextual attributes specified in each dataset. On the other hand, we do not have the ground truth information on what the contextual and behavioral attributes are in the rest of the datasets with real-world anomalies shown in Table \ref{dataset_tab}. For these datasets, the true context is assumed to be the best possible choice, i.e., the context that gives the maximum detection performance among all available contexts created from the same feature set.

\subsection{Baselines}\label{Baselines}
In our experiment, we compare WisCon against 15 methods with varying characteristics. They can be divided into three categories: (i) active learning baselines, (ii) unsupervised anomaly detection baselines, and (iii) unsupervised contextual anomaly detection baselines. The details of each method is listed below: 

\textbf{Active learning baselines:}
\begin{itemize}
    \item \textbf{AAD:} Active Anomaly Discovery (AAD) \citep{das2016incorporating,das2017incorporating} is one of the recent
methods for incorporating expert feedback into an ensemble of anomaly detectors. AAD has been implemented with different ensemble detectors such as LODA \citep{das2016incorporating} and iForest \citep{das2016incorporating}. In this work, we use iForest-AAD  as it has been shown to outperform LODA-AAD in \citep{das2017incorporating}.

\item  \textbf{Active-RF:} Active Random Forest is the active version of Random Forest classifier. It uses pool-based setting with uncertainty sampling to query instances under budget $b$.  
\item  \textbf{Active-KNN:} Active KNN is the active version of the KNN classifier. It uses pool-based setting with uncertainty sampling to query instances under budget $b$.   
\item  \textbf{Active-SVM:} Active-SVM is the active version of SVM classifier. It uses pool-based active learning with uncertainty sampling.  
\end{itemize}

\textbf{Unsupervised baselines (contextual):}
\begin{itemize}

\item  \textbf{ROCOD:} Robust contextual outlier detection  \citep{liang2016robust} (ROCOD) simultaneously considers local and global effects in outlier detection. Specically, kNN regression is used to generate a local expectation for each sample, and a ridge regression (ROCOD.RIDGE) or tree regression (ROCOD.CART) is used to produce a global expectation for each sample. Then, these two estimations are combined to generate a total expectation for the behavioral attribute value. In this work, we use ROCOD.CART since it has been shown to perform better in \citep{liang2016robust}.

\item \textbf{ConOut:} ConOut is a contextual anomaly detection algorithm that combines multiple contexts identified automatically. It measures the pairwise dependencies between attributes in a feature set and combines attributes into contexts in which highly dependent features are not present together. Then, it quantifies the outlierness of a sample in each context with a custom anomaly detector called context-incorporated outlier detection, and takes the maximum of the scores across all contexts as final outlier scores.

\item  \textbf{CAD:} Conditional Anomaly Detection (CAD) \citep{song2007conditional} is a generative approach to model the relation between context and behavior. Both the context and the behavior are modeled separately as a mixture of multiple Gaussians.  With a probabilistic mapping function, it captures how behavioral attributes are related to contextual attributes. 

\item \textbf{iForest-Con:} Contextual Isolation Forest (IF-Con) is implemented following the similar technique explained in section \ref{basedetector}, in which an anomaly detector (i.e., iForest) is trained only using behavioral attributes for each reference group, while the reference groups are estimated by XMeans using contextual attributes. 

\item \textbf{LOF-Con:} Contextual Local Outlier Factor (IF-Con) is implemented similar to iForest-Con. Instead of iForest, it uses LOF algorithm to detect anomalies in each reference group.

\item \textbf{OCSVM-Con:} Contextual One-Class SVM (IF-Con) is implemented similar to iForest-Con and LOF-Con. It uses OCSVM algorithm to detect anomalies in each reference group.
\end{itemize}

\textbf{Unsupervised baselines (non-contextual):}
\begin{itemize}
\item  \textbf{iForest:} Isolation Forest (iForest) \citep{liu2008isolation} is the isolation based tree ensemble detector. The algorithm randomly partitions the data on randomly selected features and stores this partitioning in a tree structure. The samples that travel shorter into the tree are assumed to more likely to be anomalies as they require less cuts to isolate them. 

\item \textbf{LOF:} Local Outlier Factor (LOF) \citep{breunig2000lof} is a widely used anomaly detection algorithm that compares the local density of each point to that of its neighbors. Points with significantly lower density compared to their neighbors are regarded as outliers.

\item \textbf{OC-SVM:} One class SVM \citep{scholkopf1999support} is another popular method based on the principles of support vectors.  The method projects the data to a high-dimensional feature space and tries to find a hyperplane best separating the data from the origin.

\item \textbf{LODA:} Lightweight online detector of anomalies (LODA) \citep{pevny2016loda} is an ensemble method using 1-dimensional random projections in combination with histogram-based detectors to spot anomalies in high-dimensional data. 

\item \textbf{SOD:} Subspace outlier detection (SOD) \citep{kriegel2009outlier} aims at detecting outliers that are visible in different subspaces of a high dimensional feature space. It uses shared nearest neighbors as a reference set
for each object and derives a subspace where the reference set exhibits low variance.

\item \textbf{FB:} Feature bagging (FB) \citep{lazarevic2005feature} is also an ensemble detector based on randomly sampling subspaces and combining anomaly scores from these subspaces measured by LOF algorithm.
\end{itemize}

\subsection{Experimental Setup} \label{Setup}
For all experiments reported in this section, the datasets are split into stratified training sets (70\%) and testing sets (30\%). First, we apply a hyperparameter optimization for the baselines to give them the advantage of reporting the best model performances against WisCon in each dataset. We perform a grid search over 10-fold cross-validation with 70\% of data to optimize the parameters of each method. The detailed list of parameters considered for each method can be found in Table \ref{hyperparameters}. 

As an exception, we do not apply parameter search for AAD and use the values recommended by \cite{das2017incorporating}. It has a specific optimization procedure to estimate the detector weights, and it is too computationally costly to apply this process within a grid search. Moreover, we provide true contamination rates of the datasets to all the baselines, but not to WisCon. We believe that how much a dataset is contaminated with anomalies is very difficult to know in practice, and therefore, WisCon's base detector computes anomaly scores without knowing this parameter.

Respecting the unsupervised nature of the problem, we are limiting the parameter search for WisCon. WisCon's base detector relies on iForest and XMeans algorithms, both of which include hyperparameters. The parameters of iForest are set, across all datasets, to values suggested by \cite{liu2008isolation}. XMeans algorithm automatically decides the number of clusters (i.e., reference groups) in the given contexts using Bayesian information criterion (BIC); however, it requires specifying the maximum number of clusters ($x$) as a parameter. In order to recommend a reasonable $x$ value, we test our method with $x= \{5, 10, 20\}$. In most datasets, we have not observed major effect on performance, especially for $x=10$ and $x=20$ (since XMeans will choose a lower number of clusters if necessary, i.e., the value makes no difference as long as the optimum number of clusters is smaller than $x$). However, when a larger $x$ is specified, XMeans may create clusters with very few samples. The algorithm only considers maximizing BIC, not whether resulting clusters are suitable for anomaly detection. Given that we create a separate iForest for each cluster, small clusters lead to poor models and incorrect assessments of anomalies. Therefore, we use $x=10$ as ``the maximum number of clusters'' in most datasets, unless that results in very small clusters; then, we set $x=5$. 

Another parameter for WisCon is the bias factor $\lambda$ used by the LCA sampling (see Section \ref{lca}). It decides how biased the sampling should be towards low confident anomalies, where $\lambda=0$ results in uniform sampling. We want our strategy to be heavily-biased while still preserving the opportunity of exploring new random instances from time to time. The choices for the bias factor are suggested as $0.96 \leq \lambda \leq 0.98$ in related works \citep{gama2009issues,10.5555/1083592.1083674}, therefore we use a constant $\lambda = 0.96$.

After parameter optimization, the independent test set containing 30\% of the data is used to evaluate the performances of all the methods. Each baseline is trained with the parameter values yielding the best performance in each dataset for the given performance metric. Note that for both parameter selection and evaluation, ground truth labels are only used to assert performances, not for training models. For the active learning baselines, including WisCon, samples queried for labels are selected from the training set, while test samples remain completely unseen. 

We report both the area under precision-recall curve (AUC-PR, or average precision) and the area under receiving-operating characteristics (AUC-ROC) for the evaluation of WisCon and other baselines. In each table presented throughout this section, performances are shown as the averages, together with standard deviation of scores computed in $10$ independent runs.

Considering the large differences in the magnitude of performances among datasets, we perform the Wilcoxon signed-rank test to show whether there is statistically significant difference between WisCon and another method for a specific dataset. Furthermore, we follow Demsar \citep{demvsar2006statistical} to statistically compare multiple methods over many datasets and apply Friedman test on the average ranks of the methods and Nemenyi post-hoc test. For all tests, $p < 0.05$ is considered to be significant.

For low-dimensional datasets (defined as those with the number of features $d<15$), the set of contexts that WisCon takes as an input is assumed to include all possible contexts that can be generated from the feature set. More specifically, a context is generated by dividing the feature set into two sets of attributes where the first set gives us the context and the second set is the behavior. This allows demonstrating how successfully WisCon performs given the complete knowledge of contexts in different datasets (sections \ref{Q1} and \ref{Q2}) and also analyzing how the detection performances vary among all possible contexts. However, it results in $2^d-2$ contexts, which makes the computations for higher dimensional datasets infeasible. Therefore, for datasets with $d \geq 15$, we use Principal Component Analysis (PCA) \citep{wold1987principal} to reduce the dimensionality. The number of components is set to $10$ considering that it still gives large number of contexts while managing reasonable run-time. 
There is a rich literature on feature subset selection, also for anomaly detection. Other approaches for reducing the number of considered contexts can be directly used withing WisCon. We choose PCA due to it being a simple, popular and fast method. 

Unsupervised contextual baselines (except ConOut) described in Section \ref{Baselines} expect a pre-specified context for each dataset. However, the true contexts are only known in perturbed datasets with contextual anomalies. For the other datasets, we applied an exhaustive search among all possible contexts and assume the true contexts to be the one resulting with the highest performance measured by given metric. However, such an exhaustive search is computationally expensive and can not be applied to high-dimensional datasets. Therefore, we also use PCA transformed datasets for contextual baselines to be able to determine the best possible context for these methods.




\subsection{Results Q1: Comparison of the WisCon Over Baselines} \label{Q1}

In this experiment, we compare WisCon against 15 baselines in different categories on 18 datasets using two performance metrics. Tables \ref{active_pr} and \ref{active_roc} show the performance of the proposed WisCon and active learning baselines on 18 datasets measured by AUC-PR and AUC-ROC, respectively. Each table compares WisCon against all active learning models under different budgets.

The results show that our approach performs best in majority of datasets and reports the highest average rank among all methods for all three budgets (i.e, b=\{20, 60, 100\}) across both metrics. The clear benefit of WisCon in terms of low-cost learning can be observed particularly in the performance difference between WisCon and other methods for the smallest budget of 20. Although WisCon can still mostly maintain better results compared to its competitors when using higher budgets, other methods provide notably weaker performances in many datasets under the low budget. It is especially apparent for the datasets with a large number of instances such as Synthetic 1, El Nino, Houses, and Mammography. Furthermore, WisCon is significantly better than other active baselines in detecting contextual anomalies. It significantly outperforms other methods in all perturbed datasets considering both AUC-PR and AUC-ROC results. 

AAD, a state-of-the-art active anomaly detection approach, does not show clear superiority over Active-RF for any budgets. However, both WisCon and AAD are significantly better than active classifiers among all synthetic datasets. Some real-world datasets included in the experiments are transformed from multi-class datasets, where minority classes are assumed to be the anomaly class. Active classifiers can effectively learn these minority classes and, therefore, may show high performances in some datasets. However, they cannot perform as well as anomaly detection models when detecting contextual or global anomalies similar to the synthetic examples.

We statistically compare the average ranks of the algorithms in different budgets (reported in Table 3) using the nonparametric Friedman test \cite{}. With respect to AUC-PR results, the tests return $p=\{1.26\times10^{-7},3.1 \times 10^{-3}, 0.092\}$ for $b= \{20, 60, 100\}$, respectively. Only for $b=100$ we could not reject the null hypothesis (that all methods perform equally) at a 5\% confidence-level. It clearly indicates that the performance gap among the methods decreases when the budget increases. On the other hand, we could reject the null hypothesis with $p=\{9.18\times10^{-7}, 0.014, 0.012\}$ for all three budgets when using AUC-ROC. 

\begin{figure}
\centering
\begin{subfigure}{0.7\columnwidth}
\includegraphics[width=\columnwidth]{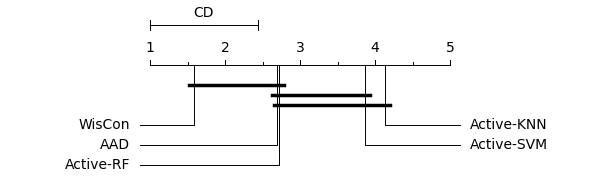}%
\caption{CD diagram B=20 (AUC-PR)}%
\label{subfiga}%
\end{subfigure}%
\begin{subfigure}{0.7\columnwidth}
\includegraphics[width=\columnwidth]{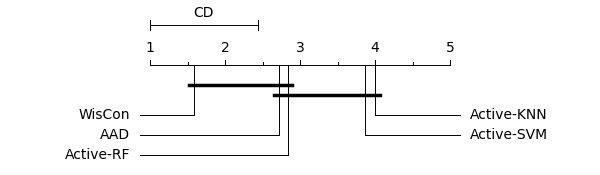}%
\caption{CD diagram B=20 (AUC-ROC)}%
\label{subfigb}%
\end{subfigure}%

\begin{subfigure}{0.7\columnwidth}
\includegraphics[width=\columnwidth]{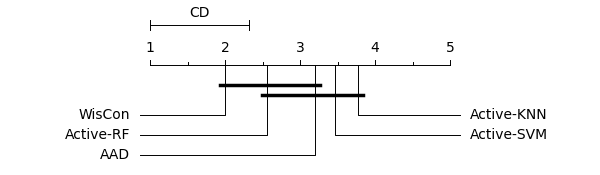}%
\caption{CD diagram B=60 (AUC-PR)}%
\label{subfigc}%
\end{subfigure}%
\begin{subfigure}{0.7\columnwidth}
\includegraphics[width=\columnwidth]{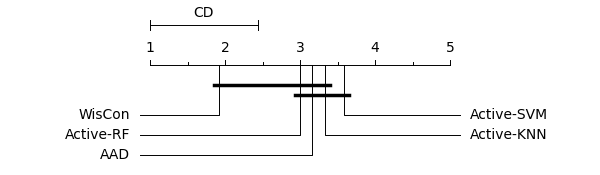}%
\caption{CD diagram B=60 (AUC-ROC)}%
\label{subfige}%
\end{subfigure}%

\begin{subfigure}{0.7\columnwidth}
\includegraphics[width=\columnwidth]{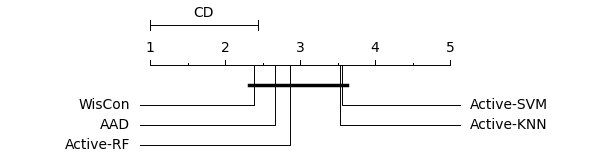}%
\caption{CD diagram B=100 (AUC-PR)}%
\label{subfigf}%
\end{subfigure}%
\begin{subfigure}{0.7\columnwidth}
\includegraphics[width=\columnwidth]{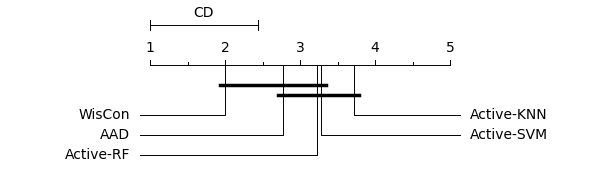}%
\caption{CD diagram B=100 (AUC-ROC)}%
\label{subfigg}%
\end{subfigure}%
\caption{The distribution of the performances (AUC-PR) in different contexts for all datasets.}
\label{cd}
\end{figure}

Next, we perform the Nemenyi post-hoc test to compare the methods in pairs while accounting for multiple testing. The test shows, for a given budget and performance metric, whether the average ranks of two methods differ by a ``critical difference'' (CD), for the 5 methods, on 18 datasets, at a significance level $\alpha = 0.05$. Fig.~\ref{cd} shows the critical difference diagrams for the three different budgets and two metrics.

In Table \ref{unsupervised_pr} and Table \ref{unsupervised_roc}, we report comparisons of WisCon (b=100) against unsupervised anomaly detection algorithms, including both contextual and non-contextual, with regards to AUC-PR and AUC-ROC. It can be seen that WisCon is able to achieve the top performance in most of datasets. Our approach ranks top in average on both metrics among state-of-the-art anomaly detectors with different properties.

As explained in Section \ref{Setup}, single-context anomaly detection methods are provided with the true context (i.e., the known true context in a perturbed dataset or the context giving the highest performance in a real dataset) for each dataset, which is mostly unknown in real-world applications. It is evident that even under such best-case scenario, a user-defined, single-context setting is inferior to the effective combination of multiple-contexts with a low budget.


ConOut is the only multi-context anomaly detector in the literature, and thus among the baselines. It uses a statistical measure to analyze dependency among different attributes and form contexts and behaviors by ensuring that highly correlated attributes are not present together. This method reduces the total number of contexts by eliminating redundant contexts ``assumed'' to create similar reference groups (i.e., subpopulations) in the ensemble. However, it cannot computationally handle very high-dimensional datasets (i.e., Arrhythmia and Optdigits) and datasets with large number of samples (i.e., El Nino). Consequently, the results of ConOut for such datasets are omitted from Tables \ref{unsupervised_pr} and \ref{unsupervised_roc}. ConOut's results for remaining datasets suggest that eliminating redundant contexts does not guarantee incorporating relevant contexts in which anomalies stand out. ConOut ranks, in average, under WisCon and also under other contextual methods (except CAD). It does not achieve adequate performances, even in perturbed datasets-- Synthetic 1, 2 and 3, that actually contain contextual anomalies. The main reason for this is that ConOut measures high correlation between all the attributes in these datasets and forms contexts eminently dissimilar to the ``true'' contexts, the ones in which the anomalies are injected. Furthermore, as an ensemble combination strategy, the maximization of anomaly scores tend to overestimate anomaloussness of an object (see \cite{aggarwal2015theoretical,zimek2014ensembles}). Given enough contexts, it is not unexpected to find a context such that a normal point appears to be an anomaly.



Another noteworthy observation is that contextual methods outperform their non-contextual equivalents in most cases, including in the non-perturbed datasets. This proves that contextual anomalies are common in different real-world public datasets and supports the need for contextual anomaly detection methods that can handle complex anomalies. Yet, most of the proposed approaches in anomaly detection today only deal with global anomalies. Furthermore, state-of-the-art contextual anomaly detection methods such a ConOut, ROCOD and CAD do not give significantly better results than iForest-Con, LOF-Con, and OCSVM-Con, the simpler and more intuitive detectors that we have implemented for these experiments.

LODA, SOD and FB are subspace methods that are designed to handle anomalies hidden in high-dimensional feature space. Unlike contextual anomaly detectors, these methods do not distinguish between contextual and behavioral attributes. As expected, they perform notably worse than WisCon in perturbed datasets with contextual anomalies, and can not outperform our approach in other datasets either. This indicates that random subspace methods can not effectively tackle contextual anomalies. 

WisCon achieves reasonably good performances on high-dimensional datasets as well. For example, it significantly outperforms all the baselines from different categories in Arrhythmia dataset, which has the highest dimensionality among all datasets. 

\begin{figure}[h!]
\begin{subfigure}{0.7\columnwidth}
\includegraphics[width=\columnwidth]{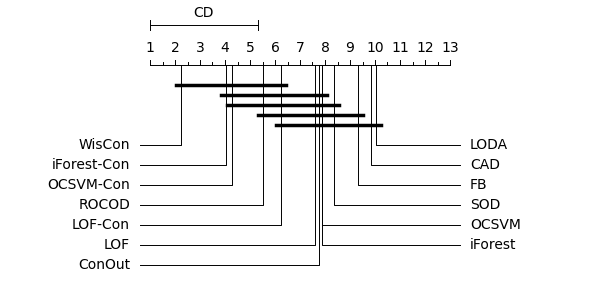}%
\caption{CD diagram (AUC-PR)}%
\label{cd_pr}%
\end{subfigure}%
\begin{subfigure}{0.7\columnwidth}
\includegraphics[width=\columnwidth]{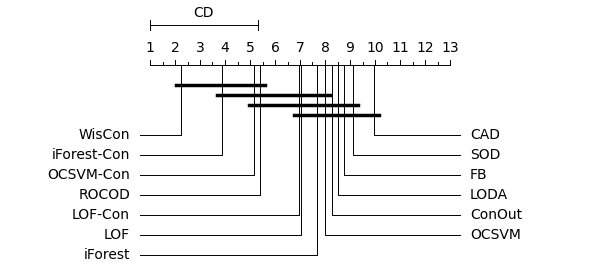}%
\caption{CD diagram (AUC-ROC)}%
\label{subfigg}%
\end{subfigure}%
\caption{The distribution of the performances (AUC-PR) in different contexts for all datasets.}
\label{cd2}
\end{figure}

According to the Friedman tests on average ranks reported in both Table \ref{unsupervised_pr} and Table \ref{unsupervised_roc}, we can conclude that methods are not statistically equivalent in terms of performances measured with AUC-PR ($p=3.33 \times 10^{-16}$) and AUC-ROC ($p=1.27 \times 10^{-12}$). The CD diagrams of WisCon and unsupervised baselines for both metrics can be observed in Fig.~\ref{cd2}. It can be seen that WisCon shows superiority over most of the competitors in this category. However, the differences between the overall performances of WisCon, iForest-Con, OCSVM-Con and ROCOD are not critical in both figures. On the other hand, WisCon shows superiority over these methods when it is compared to each of them separately. According to Wilcoxon signed-rank tests comparing two methods in each dataset, WisCon significantly outperforms any of these baselines in at least 12 out of 18 datasets for both metrics.   

In general, there is no clear winner among unsupervised general baselines in their own category. There is no significant difference between the best (i.e., LOF) and the worst (i.e, LODA or SOD) non-contextual general anomaly detection methods in either metric.

\begin{table}[t]

\caption{Performance (AUC-PR) comparisons of WisCon with active learning methods on 18 datasets using three different budgets, $b \in \{20,60,200\}$. Average ranks per budget across all datasets are given at the bottom of each subtable. The winner performances are shown in bold. The symbol $\downarrow$ $(p<0.05)$ denotes the cases that are significantly lower than the winner w.r.t. Wilcoxon signed-rank test.}

\begin{center}
\renewcommand{\arraystretch}{1.2}

\begin{tabular}{ll|ccccc}
\hline

\hline
& Dataset/Method & WisCon & AAD & Active-RF & Active-KNN & Active-SVM\\
\hline


\multirow{ 15}{*}{\rotatebox[origin=c]{90}{b=20}} &Synthetic1 & $\mathbf{0.82 \pm 0.02}$ &$0.24 \pm 0.09\downarrow$ & $0.19 \pm 0.10\downarrow$  &$0.10 \pm 0.03\downarrow$  & $0.06 \pm 0.06\downarrow$ \\
&Synthetic2 &$\mathbf{0.69 \pm 0.08}$  &$0.36 \pm 0.05\downarrow$  &$0.27 \pm 0.13\downarrow$  &$0.06 \pm 0.03\downarrow$ & $0.01 \pm 0.0\downarrow$ \\
&Synthetic3 &$\mathbf{0.86 \pm 0.04}$  &$0.12 \pm 0.05\downarrow$  &$0.09 \pm 0.05\downarrow$  & $0.02 \pm 0.01\downarrow$ & $0.01 \pm 0.0\downarrow$\\
&Synthetic4 & $\mathbf{0.99 \pm 0.0}$  &$\mathbf{0.99 \pm 0.0}$  &$0.32 \pm 0.12\downarrow$  & $0.23 \pm 0.11\downarrow$ &$0.58 \pm 0.27\downarrow$\\
&El Nino & $\mathbf{0.62 \pm 0.04}$ &$0.21 \pm 0.0\downarrow1$ &$0.08 \pm 0.05\downarrow$  &$0.11 \pm 0.01\downarrow$  & $0.10 \pm 0.03\downarrow$ \\
&Houses & $\mathbf{0.63 \pm 0.09}$ &$0.08 \pm 0.02\downarrow$ &$0.21 \pm 0.15\downarrow$  &$0.11 \pm 0.06\downarrow$  & $0.26 \pm 0.20\downarrow$ \\
&Abalone & $\mathbf{0.73 \pm 0.08}$&$0.66 \pm 0.08\downarrow$ &$0.22 \pm 0.12\downarrow$  &$0.06 \pm 0.04\downarrow$  & $0.16 \pm 0.13\downarrow$ \\
&Annthyroid  & $\mathbf{0.75 \pm 0.05}$ &$0.32 \pm 0.06\downarrow$&$0.71 \pm 0.05$  &$0.17 \pm 0.04\downarrow$  & $0.46 \pm 0.16\downarrow$ \\
&Arrhythmia &$\mathbf{0.58 \pm 0.10}$ &$0.50 \pm 0.06\downarrow$ &$0.31 \pm 0.10\downarrow$ & $0.21 \pm 0.05\downarrow$ & $0.10 \pm 0.01\downarrow$ \\
&Letters &$\mathbf{0.36 \pm 0.07}$ &$0.11 \pm 0.02\downarrow$ & $0.13 \pm 0.05\downarrow$ & $0.07 \pm 0.01\downarrow$ & $0.13 \pm 0.11\downarrow$ \\
&Mammography & $\mathbf{0.48 \pm 0.07}$ &$0.35 \pm 0.10\downarrow$ &$0.24 \pm 0.13\downarrow$  &$0.20 \pm 0.01\downarrow$  & $0.40 \pm 0.08\downarrow$ \\
&Optdigits &$0.58 \pm 0.07\downarrow$ &$0.13 \pm 0.19\downarrow$  & $\mathbf{0.99 \pm 0.0}$ & $0.31 \pm 0.11\downarrow$  & $0.79 \pm 0.38$\\
&Pendigits &$0.72 \pm 0.08\downarrow$ &$0.48 \pm 0.11\downarrow$  &$\mathbf{0.83 \pm 0.14}$  & $0.66 \pm 0.13\downarrow$ & $0.26 \pm 0.39\downarrow$ \\
&Satellite &$0.53 \pm 0.06\downarrow$   &$\mathbf{0.65 \pm 0.02}$ & $0.61 \pm 0.18$ & $0.60 \pm 0.05\downarrow$ & $0.40 \pm 0.22$ \\
&SatImage &$0.82 \pm 0.1\downarrow$  &$\mathbf{0.94 \pm 0.03}$  &$0.91 \pm 0.01\downarrow$ & $0.89 \pm 0.01\downarrow$ & $0.42 \pm 0.40$\\
&Thyroid &$0.85 \pm 0.12$ &$0.68 \pm 0.02\downarrow$  &$\mathbf{0.92 \pm 0.03}$  &$0.59 \pm 0.11\downarrow$  & $0.30 \pm 0.36\downarrow$\\
&Vowels & $\mathbf{0.63 \pm 0.02}$&$0.39 \pm 0.11\downarrow$ &$0.57 \pm 0.04\downarrow$  &$0.22 \pm 0.09\downarrow$  & $0.19 \pm 0.15\downarrow$ \\
&Yeast &$\mathbf{0.41 \pm 0.10}$  &$0.31 \pm 0.09\downarrow$  &$0.08 \pm 0.06\downarrow$ & $0.08 \pm 0.03\downarrow$ & $0.07 \pm 0.04\downarrow$  \\
\hline 
&Avg Rank &$\mathbf{1.58}$ $\mathbf{(1)}$ &$2.69$ (2) &$2.72$ (3)& $4.13$ (5)& $3.86$ (4)\\
\hline
\hline

\multirow{ 15}{*}{\rotatebox[origin=c]{90}{b=60}}&Synthetic1 & $\mathbf{0.85 \pm 0.01}$  & $0.41 \pm 0.06\downarrow$ & $0.49 \pm 0.05\downarrow$  &$0.35 \pm 0.08\downarrow$  & $0.24 \pm 0.01\downarrow$ \\
&Synthetic2 &$\mathbf{0.71 \pm 0.07}$   & $0.48 \pm 0.06\downarrow$  &$0.26 \pm 0.15\downarrow$  & $0.23 \pm 0.14\downarrow$ &$0.30 \pm 0.18\downarrow$\\
&Synthetic3 &$\mathbf{0.85 \pm 0.05}$    & $0.14 \pm 0.03\downarrow$  & $0.07 \pm 0.05\downarrow$ & $0.13 \pm 0.04\downarrow$&$0.07 \pm 0.17\downarrow$ \\
&Synthetic4 &$\mathbf{1.0 \pm 0.0}$    & $\mathbf{1.0 \pm 0.0}$  &$0.41 \pm 0.23\downarrow$  &$0.53 \pm 0.11\downarrow$& $0.79 \pm 0.21$ \\
&El Nino & $\mathbf{0.68 \pm 0.01}$  & $0.25 \pm 0.03\downarrow$ &$0.31 \pm 0.03\downarrow$  &$0.15 \pm 0.03\downarrow$  & $0.23 \pm 0.06\downarrow$ \\
&Houses & $\mathbf{0.66 \pm 0.04}$  & $0.35 \pm 0.06\downarrow$ &$0.57 \pm 0.11\downarrow$  &$0.24 \pm 0.10\downarrow$  & $0.22 \pm 0.09\downarrow$ \\
&Abalone & $\mathbf{0.79 \pm 0.07}$ & $0.64 \pm 0.12\downarrow$ &$0.14 \pm 0.13\downarrow$  &$0.09 \pm 0.13\downarrow$  & $0.08 \pm 0.13\downarrow$ \\
&Annthyroid & $\mathbf{0.77 \pm 0.04}$  & $0.35 \pm 0.06$ &$0.76 \pm 0.04\downarrow$  &$0.32 \pm 0.02\downarrow$  & $0.69 \pm 0.05\downarrow$ \\
&Arrhythmia &$\mathbf{0.60 \pm 0.09}$  & $0.30 \pm 0.07\downarrow$ &$0.30 \pm 0.08\downarrow$  & $0.15 \pm 0.04\downarrow$ & $0.20 \pm 0.12\downarrow$ \\
&Letters &$\mathbf{0.35 \pm 0.08}$  & $0.19 \pm 0.04\downarrow$ & $0.16 \pm 0.07\downarrow$ &  $0.14 \pm 0.04\downarrow$ & $0.18 \pm 0.10\downarrow$\\
&Mammography & $\mathbf{0.49 \pm 0.06}$  & $0.40 \pm 0.08$ &$0.48 \pm 0.02$  &$0.38 \pm 0.06\downarrow$  & $0.48 \pm 0.05$ \\
&Optdigits &$0.60 \pm 0.07\downarrow$  & $0.47 \pm 0.37\downarrow$  &$\mathbf{1.0 \pm 0.0}$  & $0.99 \pm 0.0$ & $0.99 \pm 0.0$\\
&Pendigits &$0.68 \pm 0.07\downarrow$   & $0.67 \pm 0.09\downarrow$  & $0.97 \pm 0.0\downarrow$ & $\mathbf{0.98 \pm 0.0}1$ & $\mathbf{0.98 \pm 0.0}$\\
&Satellite &$0.57 \pm 0.04\downarrow$   & $0.67 \pm 0.02\downarrow$ &$\mathbf{0.81 \pm 0.14}$  & $0.77 \pm 0.03$ & $0.71 \pm 0.19$ \\
&SatImage &$0.84 \pm 0.09\downarrow$  & $\mathbf{0.96 \pm 0.02}$  & $0.91 \pm 0.01$ & $0.91 \pm 0.02$ & $0.91 \pm 0.02$\\
&Thyroid &$0.86 \pm 0.07\downarrow$  & $0.83 \pm 0.06\downarrow$  &$\mathbf{0.97 \pm 0.01}$  &  $0.79 \pm 0.05\downarrow$ & $0.28 \pm 0.35\downarrow$\\
&Vowels & $0.65 \pm 0.08\downarrow$ & $0.63 \pm 0.12\downarrow$&$\mathbf{0.98 \pm 0.01}$  &$0.76 \pm 0.07\downarrow$  & $0.69 \pm 0.34\downarrow$ \\
&Yeast &$\mathbf{0.41 \pm 0.11}$  & $0.34 \pm 0.07$  &$0.06 \pm 0.02\downarrow$ & $0.03 \pm 0.0\downarrow$ & $0.06 \pm 0.01\downarrow$ \\
\hline
&Avg Rank & $\mathbf{2.0}$ $\mathbf{(1)}$& $3.19$ (3) &$2.55$ (2)& $3.77$ (5)& $3.47$ (4)\\
\hline
\hline
\multirow{ 15}{*}{\rotatebox[origin=c]{90}{b=100}}&Synthetic1 & $\mathbf{0.88 \pm 0.01}$ & $0.45 \pm 0.04\downarrow$ &$0.61 \pm 0.10\downarrow$  &$0.44 \pm 0.11\downarrow$  & $0.36 \pm 0.06\downarrow$ \\
&Synthetic2 &$\mathbf{0.72 \pm 0.06}$ & $0.53 \pm 0.05\downarrow$  &$0.20 \pm 0.12\downarrow$   & $0.31 \pm 0.08\downarrow$ &$0.33 \pm 0.16\downarrow$  \\
&Synthetic3 &$\mathbf{0.86 \pm 0.04}$ &$0.16 \pm 0.03\downarrow$  & $0.03 \pm 0.03\downarrow$   &$0.25 \pm 0.12\downarrow$  &$0.04 \pm 0.06\downarrow$  \\
&Synthetic4 &$\mathbf{1.0 \pm 0.0}$ & $\mathbf{1.0 \pm 0.0}$ &$0.54 \pm 0.24\downarrow$   &$0.84 \pm 0.18\downarrow$  &$0.79 \pm 0.20\downarrow$   \\
&El Nino & $\mathbf{0.76 \pm 0.01}$ &$0.27 \pm 0.02\downarrow$ &$0.51 \pm 0.05\downarrow$  &$0.67 \pm 0.01\downarrow$  & $0.29 \pm 0.13\downarrow$ \\
&Houses & $\mathbf{0.69 \pm 0.03}$ &$0.35 \pm 0.06\downarrow$ & $0.62 \pm 0.06\downarrow$  &$0.28 \pm 0.10\downarrow$  & $0.27 \pm 0.17\downarrow$ \\
&Abalone & $\mathbf{0.81 \pm 0.09}$&$0.62 \pm 0.17\downarrow$  &$0.12 \pm 0.14\downarrow$  &$0.07 \pm 0.09\downarrow$  & $0.09 \pm 0.16\downarrow$ \\
&Annthyroid & $\mathbf{0.80 \pm 0.04}$ &$0.44 \pm 0.08\downarrow$& $\mathbf{0.80 \pm 0.02}$  &$0.35 \pm 0.01\downarrow$  & $0.73 \pm 0.04\downarrow$ \\
&Arrhythmia &$\mathbf{0.60 \pm 0.09}$ &$0.18 \pm 0.03\downarrow$ &$0.25 \pm 0.07\downarrow$  & $0.08 \pm 0.02\downarrow$ & $0.31 \pm 0.12\downarrow$\\
&Letters &$\mathbf{0.36 \pm 0.08}$ &$0.22 \pm 0.06\downarrow$ & $0.14 \pm 0.07\downarrow$ & $0.14 \pm 0.04\downarrow$ & $0.20 \pm 0.07\downarrow$\\
&Mammography &$0.50 \pm 0.05$ &$0.50 \pm 0.05$ &$\mathbf{0.52 \pm 0.02}$  &$0.41 \pm 0.04\downarrow$  & $\mathbf{0.52 \pm 0.03}$ \\
&Optdigits &$0.62 \pm 0.08\downarrow$  & $0.80 \pm 0.21$ & $\mathbf{1.0 \pm 0.0}$ &  $\mathbf{1.0 \pm 0.0}$ & $\mathbf{1.0 \pm 0.0}$\\
&Pendigits &$0.72 \pm 0.09\downarrow$  & $0.93 \pm 0.03\downarrow$ & $\mathbf{0.99 \pm 0.0}$ & $0.98 \pm 0.0\downarrow$ & $\mathbf{0.99 \pm 0.0}$\\
&Satellite &$0.60 \pm 0.04\downarrow$   &$0.69 \pm 0.02\downarrow$  &$\mathbf{0.89 \pm 0.02}$  & $0.83 \pm 0.01\downarrow$ & $0.81 \pm 0.06\downarrow$\\
&SatImage &$0.86 \pm 0.08\downarrow$   & $\mathbf{0.98 \pm 0.0}$  &$0.96 \pm 0.01\downarrow$ & $\mathbf{0.98 \pm 0.01}$ & $0.93 \pm 0.0\downarrow$\\
&Thyroid &$0.87 \pm 0.05\downarrow$  & $0.91 \pm 0.02\downarrow$ &$\mathbf{0.97 \pm 0.01}$   &$0.82 \pm 0.06\downarrow$ & $0.35 \pm 0.28\downarrow$    \\
&Vowels & $0.65 \pm 0.08\downarrow$ &$0.74 \pm 0.07\downarrow$ &$\mathbf{0.99 \pm 0.0}$  &$0.88 \pm 0.02\downarrow$  & $0.94 \pm 0.01\downarrow$ \\
&Yeast &$\mathbf{0.41 \pm 0.10}$  &$0.39 \pm 0.06$   &$0.03 \pm 0.02\downarrow$ &  $0.03 \pm 0.01\downarrow$ & $0.04 \pm 0.02\downarrow$\\
\hline
&Avg Rank &$\mathbf{2.38}$ $\mathbf{(1)}$ &$2.66$ (2) & $2.86$ (3)& $3.52$ (4)& $3.55$ (5)\\
\hline

\hline
\end{tabular}
\label{active_pr}
\end{center}
\end{table}

\begin{table}[t]

\caption{Performance (AUC-ROC) comparisons of WisCon with active learning methods on 18 datasets using three different budgets, $b \in \{20,60,200\}$. Average ranks per budget across all datasets are given at the bottom of each subtable. The winner performances are shown in bold. The symbol $\downarrow$ $(p<0.05)$ denotes the cases that are significantly lower than the winner w.r.t. Wilcoxon signed-rank test.}

\begin{center}
\renewcommand{\arraystretch}{1.2}

\begin{tabular}{ll|ccccc}
\hline

\hline
& Dataset/Method & WisCon & AAD & Active-RF & Active-KNN & Active-SVM\\
\hline


\multirow{ 15}{*}{\rotatebox[origin=c]{90}{b=20}} &Synthetic1 & $\mathbf{0.98 \pm 0.0}$ &$0.89 \pm 0.03\downarrow$&$0.60 \pm 0.14\downarrow$  &$0.67 \pm 0.08\downarrow$  & $0.40 \pm 0.23\downarrow$ \\
&Synthetic2 &$\mathbf{0.95 \pm 0.02}$  &$0.82 \pm 0.02\downarrow$  &$0.55 \pm 0.06\downarrow$  &$0.50 \pm 0.06\downarrow$ & $0.36 \pm 0.06\downarrow$ \\
&Synthetic3 &$\mathbf{0.97 \pm 0.01}$  &$0.77 \pm 0.04\downarrow$  &$0.54 \pm 0.07\downarrow$  & $0.53 \pm 0.05\downarrow$ & $0.39 \pm 0.05\downarrow$\\
&Synthetic4 & $0.99 \pm 0.0$  & $\mathbf{1.0 \pm 0.0}$ &$0.65 \pm 0.06\downarrow$  & $0.60 \pm 0.05\downarrow$ &$0.62 \pm 0.31\downarrow$\\
&El Nino & $\mathbf{0.97 \pm 0.01}$ &$0.87 \pm 0.01\downarrow$ &$0.62 \pm 0.05\downarrow$  &$0.58 \pm 0.07\downarrow$  & $0.48 \pm 0.07\downarrow$ \\
&Houses & $\mathbf{0.97 \pm 0.0}$ &$0.81 \pm 0.04\downarrow$ &$0.81 \pm 0.02\downarrow$  & $0.77 \pm 0.04\downarrow$ & $0.56 \pm 0.18\downarrow$ \\
&Abalone & $\mathbf{0.94 \pm 0.02}$&$\mathbf{0.94 \pm 0.02}$ &$0.56 \pm 0.09\downarrow$ &$0.53 \pm 0.09\downarrow$  & $0.56 \pm 0.11\downarrow$ \\
&Annthyroid  & $\mathbf{0.97 \pm 0.0}$ &$0.79 \pm 0.05\downarrow$&$0.96 \pm 0.02$  &$0.67 \pm 0.05\downarrow$  & $0.73 \pm 0.11\downarrow$ \\
&Arrhythmia &$\mathbf{0.83 \pm 0.04}$ &$0.71 \pm 0.05\downarrow$ &$0.71 \pm 0.08\downarrow$ & $0.63 \pm 0.07\downarrow$ & $0.24 \pm 0.03\downarrow$ \\
&Letters &$\mathbf{0.73 \pm 0.04}$ &$0.61 \pm 0.07\downarrow$ & $0.61 \pm 0.06\downarrow$ & $0.54 \pm 0.05\downarrow$ & $0.47 \pm 0.12\downarrow$ \\
&Mammography & $\mathbf{0.90 \pm 0.01}$ &$0.79 \pm 0.03\downarrow$ &$0.86 \pm 0.12$  &$0.82 \pm 0.04\downarrow$  & $0.47 \pm 0.37\downarrow$ \\
&Optdigits &$0.96 \pm 0.01\downarrow$ &$0.69 \pm 0.14\downarrow$  & $\mathbf{0.99 \pm 0.0}$ & $0.96 \pm 0.02\downarrow$  & $\mathbf{0.99 \pm 0.0}$\\
&Pendigits &$0.98 \pm 0.0\downarrow$ &$0.93 \pm 0.01\downarrow$  &$0.98 \pm 0.01$  & $0.95 \pm 0.01\downarrow$ & $\mathbf{0.99 \pm 0.0}$ \\
&Satellite &$0.69 \pm 0.01$   &$0.73 \pm 0.01$ & $0.70 \pm 0.11$ & $0.73 \pm 0.02$ & $\mathbf{0.75 \pm 0.19}$  \\
&SatImage &$\mathbf{0.99 \pm 0.0}$  &$\mathbf{0.99 \pm 0.0}$  &$0.97 \pm 0.0\downarrow$ & $0.96 \pm 0.02\downarrow$ & $0.97 \pm 0.02\downarrow$\\
&Thyroid &$\mathbf{0.99 \pm 0.0}$ &$0.98 \pm 0.0$  &$\mathbf{0.99 \pm 0.0}$  &$0.95 \pm 0.02\downarrow$  & $0.97 \pm 0.03$\\
&Vowels & $\mathbf{0.94 \pm 0.01}$&$0.82 \pm 0.05\downarrow$ &$0.89 \pm 0.06\downarrow$  &$0.85 \pm 0.03\downarrow$  & $0.92 \pm 0.07$ \\
&Yeast &$\mathbf{0.71 \pm 0.06}$  &$0.68 \pm 0.04$  &$0.57 \pm 0.10\downarrow$ & $0.60 \pm 0.04\downarrow$ & $0.54 \pm 0.09\downarrow$  \\
\hline 
&Avg Rank &$\mathbf{1.58}$ $\mathbf{(1)}$ &$2.73$ (2)&$2.83$ (3) &$4.0$ (5)&$3.86$ (4) \\
\hline
\hline

\multirow{ 15}{*}{\rotatebox[origin=c]{90}{b=60}}&Synthetic1 & $\mathbf{0.98 \pm 0.0}$  & $0.92 \pm 0.01\downarrow$ & $0.62 \pm 0.10\downarrow$  &$0.81 \pm 0.04\downarrow$  & $0.61 \pm 0.13\downarrow$ \\
&Synthetic2 &$\mathbf{0.95 \pm 0.02}$   & $0.86 \pm 0.03\downarrow$  &$0.51 \pm 0.06\downarrow$  & $0.60 \pm 0.05\downarrow$ &$0.54 \pm 0.18\downarrow$\\
&Synthetic3 &$\mathbf{0.97 \pm 0.01}$    & $0.78 \pm 0.02\downarrow$  & $0.45 \pm 0.07\downarrow$ & $0.68 \pm 0.12\downarrow$&$0.40 \pm 0.04\downarrow$ \\
&Synthetic4 &$\mathbf{1.0 \pm 0.0}$    & $\mathbf{1.0 \pm 0.0}$  &$0.65 \pm 0.07\downarrow$  & $0.76 \pm 0.09\downarrow$& $0.91 \pm 0.09\downarrow$ \\
&El Nino & $\mathbf{0.98 \pm 0.0}$  & $0.85 \pm 0.01\downarrow$ & $0.68 \pm 0.06\downarrow$ &$0.68 \pm 0.02\downarrow$  & $0.47 \pm 0.08\downarrow$ \\
&Houses & $\mathbf{0.98 \pm 0.0}$  & $0.83 \pm 0.09\downarrow$ &$0.72 \pm 0.16\downarrow$  & $0.84 \pm 0.03\downarrow$ & $0.61 \pm 0.24\downarrow$ \\
&Abalone & $\mathbf{0.95 \pm 0.02}$ & $0.92 \pm 0.05$ & $0.48 \pm 0.09\downarrow$ &$0.59 \pm 0.06\downarrow$  & $0.38 \pm 0.15\downarrow$\\
&Annthyroid & $\mathbf{0.98 \pm 0.0}$  & $0.76 \pm 0.05\downarrow$ &$0.97 \pm 0.01$  &$0.73 \pm 0.03\downarrow$  &  $0.83 \pm 0.01\downarrow$\\
&Arrhythmia &$\mathbf{0.83 \pm 0.04}$  & $0.56 \pm 0.07\downarrow$ &$0.73 \pm 0.04\downarrow$  & $0.67 \pm 0.05\downarrow$ & $0.25 \pm 0.05\downarrow$ \\
&Letters &$\mathbf{0.75 \pm 0.04}$  & $0.60 \pm 0.07\downarrow$ & $0.64 \pm 0.08\downarrow$ &  $0.62 \pm 0.08\downarrow$ & $0.61 \pm 0.11\downarrow$\\
&Mammography & $\mathbf{0.90 \pm 0.01}$  & $0.82 \pm 0.02\downarrow$ &$\mathbf{0.90 \pm 0.01}$   &$0.89 \pm 0.02$  & $0.89 \pm 0.0\downarrow$ \\
&Optdigits &$0.97 \pm 0.0\downarrow$  & $0.88 \pm 0.13\downarrow$  &$\mathbf{1.0 \pm 0.0}$  & $0.99 \pm 0.0$ & $\mathbf{1.0 \pm 0.0}$\\
&Pendigits &$0.98 \pm 0.04$   & $0.98 \pm 0.01\downarrow$  & $\mathbf{0.99 \pm 0.0}$ & $\mathbf{0.99 \pm 0.0}$ & $\mathbf{0.99 \pm 0.0}$\\
&Satellite &$0.69 \pm 0.01\downarrow$   & $0.74 \pm 0.02\downarrow$ &$\mathbf{0.91 \pm 0.03}$  & $0.84 \pm 0.04\downarrow$ & $0.82 \pm 0.02\downarrow$ \\
&SatImage & $\mathbf{0.99 \pm 0.0}$ & $\mathbf{0.99 \pm 0.0}$  & $0.97 \pm 0.01\downarrow$ & $0.96 \pm 0.01\downarrow$ & $0.98 \pm 0.0\downarrow$\\
&Thyroid &$\mathbf{0.99 \pm 0.0}$  & $\mathbf{0.99 \pm 0.0}$  &$\mathbf{0.99 \pm 0.0}$  &  $0.97 \pm 0.01\downarrow$ & $\mathbf{0.99 \pm 0.0}$\\
&Vowels & $0.95 \pm 0.02\downarrow$ & $0.92 \pm 0.04\downarrow$&$0.97 \pm 0.02\downarrow$  &$0.96 \pm 0.02\downarrow$  & $\mathbf{0.99 \pm 0.0}$ \\
&Yeast &$\mathbf{0.72 \pm 0.06}$  & $0.71 \pm 0.06$  &$0.55 \pm 0.06$ & $0.55 \pm 0.08$ & $0.55 \pm 0.06$ \\
\hline
&Avg Rank & $\mathbf{1.91}$ $\mathbf{(1)}$& $3.16$ (3)& $3.0$ (2)& $3.33$ (4)& $3.58$ (5)\\
\hline
\hline
\multirow{ 15}{*}{\rotatebox[origin=c]{90}{b=100}}&Synthetic1 & $\mathbf{0.99 \pm 0.0}$ & $0.92 \pm 0.01$ &$0.65 \pm 0.10$  &$0.82 \pm 0.06$  & $0.66 \pm 0.14$ \\
&Synthetic2 &$\mathbf{0.96 \pm 0.02}$ & $0.88 \pm 0.03\downarrow$  &$0.50 \pm 0.08\downarrow$   & $0.62 \pm 0.07\downarrow$ &$0.74 \pm 0.15\downarrow$  \\
&Synthetic3 &$\mathbf{0.98 \pm 0.01}$ &$0.80 \pm 0.04\downarrow$  & $0.42 \pm 0.07\downarrow$   &$0.76 \pm 0.08\downarrow$  &$0.41 \pm 0.05\downarrow$  \\
&Synthetic4 &$\mathbf{1.0 \pm 0.0}$ & $\mathbf{1.0 \pm 0.0}$ &$0.69 \pm 0.07\downarrow$   & $0.82 \pm 0.12\downarrow$ &$0.89 \pm 0.10\downarrow$   \\
&El Nino & $\mathbf{0.98 \pm 0.0}$ &$0.86 \pm 0.01\downarrow$ &$0.71 \pm 0.10\downarrow$  &$0.71 \pm 0.06\downarrow$  & $0.50 \pm 0.07\downarrow$ \\
&Houses & $\mathbf{0.98 \pm 0.0}$ &$0.88 \pm 0.02\downarrow$ & $0.59 \pm 0.22\downarrow$  &$0.86 \pm 0.03\downarrow$ & $0.72 \pm 0.07\downarrow$ \\
&Abalone & $\mathbf{0.95 \pm 0.02}$&$0.93 \pm 0.05$  &$0.58 \pm 0.09\downarrow$  &$0.55 \pm 0.07\downarrow$  &  $0.41 \pm 0.07\downarrow$ \\
&Annthyroid & $\mathbf{0.98 \pm 0.0}$ &$0.83 \pm 0.05\downarrow$& $\mathbf{0.98 \pm 0.02}$  &$0.74 \pm 0.01\downarrow$  & $0.86 \pm 0.11\downarrow$ \\
&Arrhythmia &$\mathbf{0.83 \pm 0.04}$ &$0.43 \pm 0.05\downarrow$ &$0.75 \pm 0.05\downarrow$  & $0.59 \pm 0.08\downarrow$ & $0.31 \pm 0.02\downarrow$\\
&Letters &$\mathbf{0.76 \pm 0.04}$ &$0.69 \pm 0.06\downarrow$ & $0.74 \pm 0.04$ & $0.67 \pm 0.05\downarrow$ & $0.67 \pm 0.04\downarrow$\\
&Mammography &$\mathbf{0.91 \pm 0.01}$ &$0.85 \pm 0.03\downarrow$ &$0.90 \pm 0.02$  &$0.87 \pm 0.02\downarrow$  & $0.89 \pm 0.05$ \\
&Optdigits &$0.97 \pm 0.0\downarrow$  & $0.99 \pm 0.03$ & $\mathbf{1.0 \pm 0.0}$ &  $\mathbf{1.0 \pm 0.0}$ & $\mathbf{1.0 \pm 0.0}$\\
&Pendigits &$0.99 \pm 0.0$  & $0.99 \pm 0.0$ & $0.99 \pm 0.0$ & $0.99 \pm 0.0$ & $\mathbf{1.0 \pm 0.0}$\\
&Satellite &$0.70 \pm 0.0\downarrow$   &$0.75 \pm 0.01\downarrow$  &$\mathbf{0.91 \pm 0.04}$  & $\mathbf{0.91 \pm 0.04}$ & $0.85 \pm 0.0\downarrow$\\
&SatImage &$\mathbf{0.99 \pm 0.0}$   & $\mathbf{0.99 \pm 0.0}$  &$0.98 \pm 0.01$ & $0.96 \pm 0.02\downarrow$ & $0.98 \pm 0.01$\\
&Thyroid &$\mathbf{0.99 \pm 0.0}$   & $\mathbf{0.99 \pm 0.0}$  &$\mathbf{0.99 \pm 0.0}$   &$0.98 \pm 0.0$ & $\mathbf{0.99 \pm 0.0}$    \\
&Vowels & $0.95 \pm 0.01\downarrow$ &$0.95 \pm 0.02\downarrow$ &$0.99 \pm 0.0$  &$0.96 \pm 0.06$  & $\mathbf{1.0 \pm 0.0}$ \\
&Yeast &$0.72 \pm 0.06$  &$\mathbf{0.75 \pm 0.06}$   &$0.54 \pm 0.09\downarrow$ &  $0.59 \pm 0.04\downarrow$ & $0.57 \pm 0.07\downarrow$\\
\hline
&Avg Rank & $\mathbf{2.0}$ $\mathbf{(1)}$& $2.77$ (2)&$3.22$ (3)&$3.72$ (5)& $3.27$ (4)\\
\hline

\hline
\end{tabular}
\label{active_roc}
\end{center}
\end{table}

\begin{landscape}
\begin{table}[t]
\caption{Performance (AUC-PR) comparisons of WisCon with different types of unsupervised baselines on 18 datasets. Average ranks across all datasets are given at the bottom. The winner performances are shown in bold. The symbol $\downarrow$ $(p<0.05)$ denotes the cases that are significantly lower than the winner w.r.t. Wilcoxon signed-rank test.}

\begin{center}
\setlength{\tabcolsep}{0.3em}
\renewcommand{\arraystretch}{1.2}
\hspace*{-6cm}
\begin{tabular}{l|ccccccccccccc}
\hline

\hline
Dataset & WisCon & ConOut  & ROCOD & CAD & IF-Con &LOF-Con &OCSVM-Con &IF &LOF & OCVM &LODA &SOD &FB\\
\hline


Synthetic1 & $\mathbf{0.88 \pm 0.01}$ &$0.16 \pm 0.03\downarrow$ &$0.76 \pm 0.01\downarrow$ &$0.35 \pm 0.18\downarrow$  &$0.81 \pm 0.0\downarrow$ &$0.78 \pm 0.02\downarrow$ &$0.67 \pm 0.05\downarrow$ & $0.47 \pm 0.03\downarrow$  &$0.77 \pm 0.0\downarrow$ & $0.52 \pm 0.05\downarrow$ & $0.12 \pm 0.02\downarrow$ & $0.13 \pm 0.03\downarrow$& $0.20 \pm 0.01\downarrow$ \\
Synthetic2 &$\mathbf{0.72 \pm 0.06}$  &$0.12 \pm 0.03\downarrow$   &$0.61 \pm 0.08\downarrow$  &$0.35 \pm 0.15\downarrow$  &$0.50 \pm 0.07\downarrow$  &$0.55 \pm 0.10\downarrow$ &$0.50 \pm 0.12\downarrow$ & $0.10 \pm 0.04\downarrow$&$0.61 \pm 0.08\downarrow$ &$0.40 \pm 0.08\downarrow$  & $0.06 \pm 0.02\downarrow$ &$0.14 \pm 0.05\downarrow$  &$0.42 \pm 0.06\downarrow$   \\
Synthetic3 &$0.86 \pm 0.04\downarrow$  & $0.25 \pm 0.04\downarrow$    & $0.81 \pm 0.04\downarrow$  &$0.03 \pm 0.02\downarrow$  & $0.67 \pm 0.05\downarrow$  &$0.92 \pm 0.02\downarrow$ &$\mathbf{0.94 \pm 0.03}$ & $0.08 \pm 0.02\downarrow$ & $0.64 \pm 0.08\downarrow$ & $0.01 \pm 0.0\downarrow$ & $0.06 \pm 0.04\downarrow$ & $0.15 \pm 0.03\downarrow$ & $0.46 \pm 0.05\downarrow$ \\
Synthetic4 &$\mathbf{1.0 \pm 0.0}$ &$\mathbf{1.0 \pm 0.0}$ & $0.76 \pm 0.27\downarrow$   &$0.78 \pm 0.26\downarrow$ &$0.96 \pm 0.06$   &$0.75 \pm 0.16\downarrow$ & $0.77 \pm 0.16\downarrow$& $\mathbf{1.0 \pm 0.0}$ & $\mathbf{1.0 \pm 0.0}$ & $\mathbf{1.0 \pm 0.0}$ & $\mathbf{1.0 \pm 0.0}$ &$0.99 \pm 0.01$ & $0.87 \pm 0.07\downarrow$ \\
El Nino & $\mathbf{0.76 \pm 0.01}$& -  &$0.63 \pm 0.01\downarrow$  &$0.42 \pm 0.02\downarrow$  & $0.62 \pm 0.05\downarrow$ & $0.71 \pm 0.01\downarrow$ & $0.74 \pm 0.0\downarrow$ & $0.23 \pm 0.01\downarrow$ & $0.61 \pm 0.01\downarrow$ & $0.23 \pm 0.01\downarrow$ &  $0.17 \pm 0.03\downarrow$ &$0.43 \pm 0.01\downarrow$ &$0.58 \pm 0.02\downarrow$   \\
Houses & $\mathbf{0.69 \pm 0.03}$ &$0.21 \pm 0.02$ &$0.68 \pm 0.01$  &$0.19 \pm 0.10\downarrow$  & $0.63 \pm 0.06\downarrow$ & $0.17 \pm 0.02\downarrow$ & $0.55 \pm 0.04\downarrow$  &$0.06 \pm 0.01\downarrow$ &  $0.13 \pm 0.01\downarrow$ & $0.10 \pm 0.02\downarrow$ & $0.05 \pm 0.01\downarrow$ & $0.08 \pm 0.01\downarrow$    &  $0.08 \pm 0.11\downarrow$ \\
Abalone & $\mathbf{0.81 \pm 0.09}$ &$0.52 \pm 0.12\downarrow$ &$0.23 \pm 0.01\downarrow$  &$0.09 \pm 0.11\downarrow$  & $0.66 \pm 0.11\downarrow$ & $0.57 \pm 0.15\downarrow$ & $0.47 \pm 0.24\downarrow$ & $0.54 \pm 0.17\downarrow$ & $0.27 \pm 0.06\downarrow$ & $0.49 \pm 0.13\downarrow$ & $0.07 \pm 0.07\downarrow$ &  $0.59 \pm 0.12\downarrow$ & $0.38 \pm 0.09\downarrow$  \\
Annthyroid & $0.80 \pm 0.04$& $0.32 \pm 0.02\downarrow$&$0.58 \pm 0.03\downarrow$  &$0.53 \pm 0.06\downarrow$  & $0.80 \pm 0.02\downarrow$ & $0.72 \pm 0.03\downarrow$ & $\mathbf{0.82 \pm 0.03}$ & $0.34 \pm 0.04\downarrow$ & $0.26 \pm 0.04\downarrow$ &  $0.18 \pm 0.01\downarrow$ & $0.17 \pm 0.03\downarrow$ & $0.41 \pm 0.02\downarrow$ &  $0.09 \pm 0.0\downarrow$   \\
Arrhythmia &$\mathbf{0.60 \pm 0.09}$ & - &$0.45 \pm 0.13\downarrow$  &$0.19 \pm 0.04\downarrow$ &$0.54 \pm 0.08$  &$0.57 \pm 0.07$ &$0.47 \pm 0.06\downarrow$ & $0.47 \pm 0.11\downarrow$ &  $0.47 \pm 0.08\downarrow$ & $0.14 \pm 0.0\downarrow$ & $0.47 \pm 0.10\downarrow$ & $0.46 \pm 0.09\downarrow$ & $0.49 \pm 0.06\downarrow$     \\

Letters &$0.36 \pm 0.08\downarrow$ &$0.19 \pm 0.03\downarrow$ &$0.22 \pm 0.04\downarrow$  &$0.07 \pm 0.01\downarrow$ &$0.33 \pm 0.08\downarrow$ &$0.31 \pm 0.05\downarrow$  &$0.34 \pm 0.07\downarrow$ & $0.09 \pm 0.01\downarrow$ & $0.33 \pm 0.04\downarrow$ &$\mathbf{0.50 \pm 0.12}$  & $0.07 \pm 0.01\downarrow$ & $0.31 \pm 0.05\downarrow$ & $0.34 \pm 0.14\downarrow$ \\
Mammog & $\mathbf{0.50 \pm 0.05}$&$0.21 \pm 0.06\downarrow$ &$0.39 \pm 0.02\downarrow$  &$0.14 \pm 0.12\downarrow$  & $0.45 \pm 0.03\downarrow$ & $0.02 \pm 0.0\downarrow$ & $0.47 \pm 0.05$ & $0.23 \pm 0.04\downarrow$ & $0.18 \pm 0.03\downarrow$ & $0.24 \pm 0.03\downarrow$ & $0.18 \pm 0.08\downarrow$ & $0.18 \pm 0.04\downarrow$ & $0.04 \pm 0.0\downarrow$\\
Optdigits &$0.62 \pm 0.08\downarrow$ & - &$\mathbf{0.76 \pm 0.03}$  &$0.33 \pm 0.27\downarrow$ &$0.40 \pm 0.05\downarrow$ &$0.30 \pm 0.02\downarrow$  &$0.50 \pm 0.05\downarrow$ & $0.05 \pm 0.01\downarrow$ & $0.07 \pm 0.02\downarrow$ & $0.02 \pm 0.0\downarrow$ &$0.06 \pm 0.05\downarrow$ & $0.05 \pm 0.0\downarrow$ &  $0.07 \pm 0.02\downarrow$\\
Pendigits &$\mathbf{0.72 \pm 0.09}$ &$0.34 \pm 0.06$  &$0.52 \pm 0.05\downarrow$   &$0.25 \pm 0.16\downarrow$  &$0.41 \pm 0.07\downarrow$ &$0.10 \pm 0.04\downarrow$ & $0.52 \pm 0.18\downarrow$& $0.31 \pm 0.05\downarrow$ & $0.03 \pm 0.0\downarrow$ & $0.32 \pm 0.03\downarrow$ & $0.26 \pm 0.09\downarrow$ & $0.27 \pm 0.03\downarrow$ & $0.04 \pm 0.01\downarrow$  \\
Satellite &$0.60 \pm 0.04\downarrow$ &$\mathbf{0.68 \pm 0.01}$  &$0.61 \pm 0.01\downarrow$  &$0.51 \pm 0.05\downarrow$ & $0.62 \pm 0.01\downarrow$ &$0.56 \pm 0.01\downarrow$  &$\mathbf{0.68 \pm 0.03}$ & $0.63 \pm 0.01\downarrow$ & $0.38 \pm 0.01\downarrow$ & $0.51 \pm 0.01\downarrow$& $0.63 \pm 0.01\downarrow$ & $0.54 \pm 0.02\downarrow$ & $0.37 \pm 0.0\downarrow$ \\
SatImage & $0.86 \pm 0.08\downarrow$ & $0.93 \pm 0.04\downarrow$  & $0.96 \pm 0.02\downarrow$ &$0.12 \pm 0.06\downarrow$ & $0.90 \pm 0.05\downarrow$ &$0.32 \pm 0.0\downarrow$  &$0.96 \pm 0.03$ & $0.92 \pm 0.03\downarrow$ & $0.23 \pm 0.03\downarrow$ & $\mathbf{0.97 \pm 0.02}$ & $0.91 \pm 0.04\downarrow$ &$0.90 \pm 0.02\downarrow$  &$0.03 \pm 0.02\downarrow$ \\
Thyroid &$\mathbf{0.87 \pm 0.05}$ & $0.63 \pm 0.05\downarrow$ &$0.59 \pm 0.08\downarrow$  &$0.11 \pm 0.02\downarrow$ & $0.80 \pm 0.03\downarrow$ &$0.69 \pm 0.08\downarrow$  &$0.48 \pm 0.09\downarrow$ & $0.66 \pm 0.09\downarrow$ & $0.24 \pm 0.03\downarrow$ &$0.15 \pm 0.03\downarrow$ & $0.25 \pm 0.08\downarrow$ & $0.55 \pm 0.13\downarrow$ & $0.02 \pm 0.0\downarrow$\\
Vowels & $\mathbf{0.65 \pm 0.08}$ &$0.17 \pm 0.08\downarrow$ &$0.42 \pm 0.04\downarrow$  &$0.12 \pm 0.04\downarrow$  & $0.48 \pm 0.07\downarrow$ & $0.53 \pm 0.10\downarrow$ & $0.43 \pm 0.09\downarrow$  & $0.18 \pm 0.07\downarrow$ & $0.41 \pm 0.10\downarrow$ & $0.54 \pm 0.10\downarrow$ & $0.06 \pm 0.03\downarrow$ & $0.34 \pm 0.09\downarrow$ & $0.35 \pm 0.12\downarrow$\\
Yeast &$\mathbf{0.41 \pm 0.10}$ &  $0.28 \pm 0.10\downarrow$&  $0.28 \pm 0.07\downarrow$& $0.23 \pm 0.06\downarrow$ & $0.38 \pm 0.09$ &$0.22 \pm 0.08\downarrow$  &$0.28 \pm 0.09\downarrow$ & $0.21 \pm 0.06\downarrow$ & $0.26 \pm 0.12\downarrow$ &  $0.19 \pm 0.05\downarrow$ & $0.16 \pm 0.10\downarrow$ & $0.22 \pm 0.08\downarrow$ & $0.16 \pm 0.06\downarrow$  \\
\hline 
Avg Rank &$\mathbf{2.22}$ $\mathbf{(1)}$ & $7.77$ (7)  & $5.52$ (4)& $9.83$ (11) & $4.05$ (2) & $6.22$ (5)& $4.27$ (3)& $7.88$ (8)& $7.58$ (6)& $7.88$ (8)& $10.02$ (12)& $8.36$ (9)& $9.33$ (10)\\
\hline
\end{tabular}
\label{unsupervised_pr}
\end{center}
\end{table}
\end{landscape}

\begin{landscape}
\begin{table}[t]
\caption{Performance (AUC-ROC) comparisons of WisCon with different types of unsupervised baselines on 18 datasets. Average ranks across all datasets are given at the bottom. The winner performances are shown in bold. The symbol $\downarrow$ $(p<0.05)$ denotes the cases that are significantly lower than the winner w.r.t. Wilcoxon signed-rank test.}

\begin{center}
\setlength{\tabcolsep}{0.3em}
\renewcommand{\arraystretch}{1.2}
\hspace*{-6.8cm}
\begin{tabular}{l|cccccccccccccc}
\hline

\hline
Dataset & WisCon & ConOut  & ROCOD & CAD & IF-Con &LOF-Con &OCSVM-Con &IF &LOF & OCVM &LODA &SOD &FB\\
\hline


Synthetic1 & $\mathbf{0.99 \pm 0.0}$& $0.85 \pm 0.01\downarrow$   &$\mathbf{0.99 \pm 0.0}$ &$0.77 \pm 0.20\downarrow$  &$0.98 \pm 0.0\downarrow$ &$0.92 \pm 0.0\downarrow$ &$\mathbf{0.99 \pm 0.0}$ & $0.89 \pm 0.02\downarrow$  &$0.96 \pm 0.0\downarrow$ & $0.95 \pm 0.0\downarrow$ & $0.88 \pm 0.02\downarrow$ & $0.72 \pm 0.02\downarrow$& $0.93 \pm 0.01\downarrow$ \\
Synthetic2 &$\mathbf{0.96 \pm 0.02}$ & $0.79 \pm 0.02\downarrow$     &$0.94 \pm 0.0\downarrow$  &$0.81 \pm 0.12\downarrow$  &$0.94 \pm 0.01\downarrow$  &$0.93 \pm 0.02\downarrow$ & $0.94 \pm 0.02\downarrow$&$0.82 \pm 0.03\downarrow$  &$0.93 \pm 0.01\downarrow$  &$0.93 \pm 0.01\downarrow$ &$0.76 \pm 0.04\downarrow$ & $0.78 \pm 0.04\downarrow$ &$0.91 \pm 0.01\downarrow$     \\
Synthetic3 &$0.98 \pm 0.01$& $0.85 \pm 0.02\downarrow$  & $0.98 \pm 0.0\downarrow$  &$0.57 \pm 0.04\downarrow$  & $0.97 \pm 0.01\downarrow$  &$0.98 \pm 0.01$ &$\mathbf{0.99 \pm 0.0}$ & $0.79 \pm 0.04\downarrow$ & $0.95 \pm 0.01\downarrow$ &$0.50 \pm 0.0\downarrow$ & $0.72 \pm 0.05\downarrow$ & $0.65 \pm 0.04\downarrow$ & $0.93 \pm 0.0\downarrow$ \\
Synthetic4 &$\mathbf{1.0 \pm 0.0}$ &$\mathbf{1.0 \pm 0.0}$  &$0.98 \pm 0.0\downarrow$  &$0.70 \pm 0.18\downarrow$  &$0.98 \pm 0.03\downarrow$   &$0.83 \pm 0.14\downarrow$ &$0.96 \pm 0.01\downarrow$ & $\mathbf{1.0 \pm 0.0}$ & $\mathbf{1.0 \pm 0.0}$ & $\mathbf{1.0 \pm 0.0}$ & $\mathbf{1.0 \pm 0.0}$ &$0.99 \pm 0.01$ & $0.97 \pm 0.04\downarrow$ \\
El Nino & $\mathbf{0.98 \pm 0.0}$ & -  &$0.96 \pm 0.0\downarrow$  &$0.78 \pm 0.5\downarrow$  & $0.96 \pm 0.0\downarrow$ & $0.91 \pm 0.0\downarrow$ &$0.93 \pm 0.0\downarrow$  & $0.83 \pm 0.01\downarrow$ & $0.97 \pm 0.0\downarrow$ & $0.96 \pm 0.01\downarrow$ &  $0.82 \pm 0.05\downarrow$ & $0.92 \pm 0.01\downarrow$ & $0.97 \pm 0.0\downarrow$  \\
Houses & $\mathbf{0.98 \pm 0.0}$ & $0.83 \pm 0.01\downarrow$ &$\mathbf{0.98 \pm 0.0}$  & $0.87 \pm 0.03\downarrow$ & $0.96 \pm 0.02\downarrow$ & $0.93 \pm 0.01\downarrow$ & $0.95 \pm 0.0\downarrow$ &$0.75 \pm 0.02\downarrow$ &  $0.71 \pm 0.01\downarrow$ & $0.93 \pm 0.0\downarrow$ & $0.77 \pm 0.08\downarrow$ & $0.70 \pm 0.03\downarrow$    &  $0.75 \pm 0.01\downarrow$ \\
Abalone & $\mathbf{0.95 \pm 0.02}$ & $0.93 \pm 0.01$ &$0.84 \pm 0.05\downarrow$  &$0.53 \pm 0.14\downarrow$  & $0.94 \pm 0.04$ & $0.94 \pm 0.02$ & $0.94 \pm 0.03$ & $0.84 \pm 0.08\downarrow$ & $0.89 \pm 0.03\downarrow$ & $0.78 \pm 0.05\downarrow$ & $0.79 \pm 0.06\downarrow$ &  $0.88 \pm 0.06\downarrow$ & $0.90 \pm 0.05\downarrow$  \\
Annthyroid & $\mathbf{0.98 \pm 0.0}$ & $0.81 \pm 0.01\downarrow$& $0.94 \pm 0.0\downarrow$  &$0.96 \pm 0.01\downarrow$  & $0.97 \pm 0.0$ & $0.93 \pm 0.01\downarrow$ & $0.97 \pm 0.0$& $0.82 \pm 0.03\downarrow$ & $0.66 \pm 0.01\downarrow$ &  $0.61 \pm 0.01\downarrow$ & $0.64 \pm 0.03\downarrow$ & $0.73 \pm 0.03\downarrow$ &  $0.47 \pm 0.0\downarrow$   \\
Arrhythmia &$\mathbf{0.83 \pm 0.04}$& -  &$0.63 \pm 0.07\downarrow$  &$0.51 \pm 0.04\downarrow$ &$0.80 \pm 0.02\downarrow$  &$0.79 \pm 0.05\downarrow$ &$0.76 \pm 0.04\downarrow$ & $0.77 \pm 0.04\downarrow$ &  $0.77 \pm 0.04\downarrow$ & $0.50 \pm 0.0\downarrow$& $0.76 \pm 0.04\downarrow$ & $0.74 \pm 0.03\downarrow$ & $0.78 \pm 0.03\downarrow$     \\
Letters &$0.76 \pm 0.04\downarrow$ &$0.58 \pm 0.01\downarrow$ &$0.70 \pm 0.03\downarrow$  &$0.49 \pm 0.08\downarrow$ &$0.76 \pm 0.04\downarrow$ &$0.79 \pm 0.02\downarrow$  &$0.81 \pm 0.03\downarrow$ & $0.59 \pm 0.05\downarrow$ & $\mathbf{0.87 \pm 0.02}$ & $0.67 \pm 0.05\downarrow$& $0.50 \pm 0.03\downarrow$ & $0.86 \pm 0.01$ & $0.83 \pm 0.09$ \\
Mammog & $\mathbf{0.91 \pm 0.01}$& $0.84 \pm 0.02\downarrow$  &$0.86 \pm 0.02\downarrow$  &$0.82 \pm 0.03\downarrow$  & $0.88 \pm 0.0\downarrow$ & $0.79 \pm 0.03\downarrow$ & $0.33 \pm 0.02\downarrow$ & $0.85 \pm 0.01\downarrow$ & $0.49 \pm 0.0\downarrow$ & $0.56 \pm 0.01\downarrow$ & $0.69 \pm 0.14\downarrow$ & $0.84 \pm 0.03\downarrow$ & $0.49 \pm 0.09\downarrow$\\
Optdigits &$\mathbf{0.98 \pm 0.0}$& -  &$0.97 \pm 0.01$  &$0.83 \pm 0.23\downarrow$ &$0.95 \pm 0.0\downarrow$ &$0.95 \pm 0.0\downarrow$  &$0.96 \pm 0.0\downarrow$  & $0.67 \pm 0.02\downarrow$ & $0.59 \pm 0.05\downarrow$ &$0.40 \pm 0.01\downarrow$ &$0.47 \pm 0.04\downarrow$ & $0.56 \pm 0.09\downarrow$ &  $0.59 \pm 0.03\downarrow$\\
Pendigits &$\mathbf{0.99 \pm 0.0}$& $0.96 \pm 0.0\downarrow$ & $0.98 \pm 0.0\downarrow$   &$0.89 \pm 0.04\downarrow$  &$0.96 \pm 0.0\downarrow$ &$0.77 \pm 0.06\downarrow$ & $0.81 \pm 0.01\downarrow$ & $0.95 \pm 0.0\downarrow$ & $0.59 \pm 0.02\downarrow$ & $0.96 \pm 0.0\downarrow$ & $0.86 \pm 0.07\downarrow$ & $0.84 \pm 0.02\downarrow$ & $0.51 \pm 0.04\downarrow$  \\
Satellite &$0.70 \pm 0.0\downarrow$ & $0.71 \pm 0.01\downarrow$  &$0.74 \pm 0.01\downarrow$ &$0.69 \pm 0.06\downarrow$ & $0.73 \pm 0.01\downarrow$ &$0.70 \pm 0.01\downarrow$  &$\mathbf{0.79 \pm 0.0}$  & $0.72 \pm 0.02\downarrow$ & $0.59 \pm 0.01\downarrow$ & $0.67 \pm 0.0\downarrow$& $0.74 \pm 0.01\downarrow$ & $0.67 \pm 0.01\downarrow$ & $0.56 \pm 0.01\downarrow$ \\
SatImage & $\mathbf{0.99 \pm 0.0}$& $\mathbf{0.99 \pm 0.0}$   &$0.89 \pm 0.06\downarrow$  &$0.89 \pm 0.02\downarrow$ & $\mathbf{0.99 \pm 0.0}$ &$0.47 \pm 0.01\downarrow$  & $\mathbf{0.99 \pm 0.01}$ & $0.98 \pm 0.01$ & $0.97 \pm 0.0\downarrow$ & $\mathbf{0.99 \pm 0.0}$  & $\mathbf{0.99 \pm 0.0}$  &$0.90 \pm 0.02\downarrow$  &$0.53 \pm 0.04\downarrow$ \\
Thyroid &$\mathbf{0.99 \pm 0.0}$ & $0.95 \pm 0.01\downarrow$ &$0.95 \pm 0.01\downarrow$  &$0.77 \pm 0.24\downarrow$ & $\mathbf{0.99 \pm 0.0}$ &$0.97 \pm 0.0\downarrow$  &$0.95 \pm 0.03\downarrow$  & $\mathbf{0.99 \pm 0.0}$ & $0.95 \pm 0.01\downarrow$ &$0.79 \pm 0.03\downarrow$ & $0.95 \pm 0.01\downarrow$ & $0.92 \pm 0.04\downarrow$ & $0.47 \pm 0.0\downarrow$\\
Vowels & $\mathbf{0.95 \pm 0.01}$ & $0.77 \pm 0.03\downarrow$ &$0.91 \pm 0.03\downarrow$  &$0.86 \pm 0.08\downarrow$  & $0.93 \pm 0.05$ & $0.92 \pm 0.0\downarrow$ & $0.92 \pm 0.02$   & $0.67 \pm 0.07\downarrow$ & $0.93 \pm 0.01\downarrow$ & $0.92 \pm 0.03\downarrow$ & $0.67 \pm 0.05\downarrow$ & $0.88 \pm 0.02\downarrow$ & $0.93 \pm 0.01\downarrow$\\
Yeast &$\mathbf{0.72 \pm 0.06}$ &  $0.63 \pm 0.04\downarrow$&  $0.63 \pm 0.06\downarrow$& $0.60 \pm 0.08\downarrow$ & $0.70 \pm 0.05$ & $0.60 \pm 0.04\downarrow$ &$0.70 \pm 0.06$  & $0.59 \pm 0.07\downarrow$ & $0.68 \pm 0.05$ &  $0.71 \pm 0.05$ & $\mathbf{0.72 \pm 0.07}$ & $0.61 \pm 0.08\downarrow$ & $0.56 \pm 0.01\downarrow$  \\
\hline 
Avg Rank &$\mathbf{2.25}$ $\mathbf{(1)}$ & $8.27$ (9) & $5.41$ (4) & $9.94$ (13)& $3.88$ (2) &$6.94$ (6)& $5.16$ (3)& $7.69$ (7) & $7.02$ (5)& $8.00$ (8)& $8.50$ (10) & $9.11$ (12) & $8.77$ (11)\\
\hline
\end{tabular}
\label{unsupervised_roc}
\end{center}
\end{table}
\end{landscape}

\begin{figure*}%
\centering
\begin{subfigure}{0.35\columnwidth}
\includegraphics[width=\columnwidth]{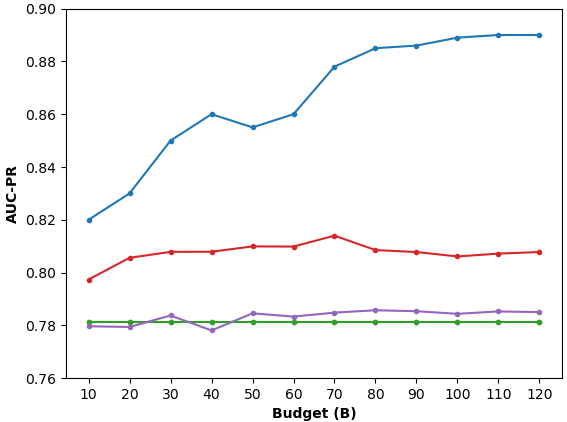}%
\caption{Synthetic Dataset}%
\label{subfiga}%
\end{subfigure}%
\begin{subfigure}{0.35\columnwidth}
\includegraphics[width=\columnwidth]{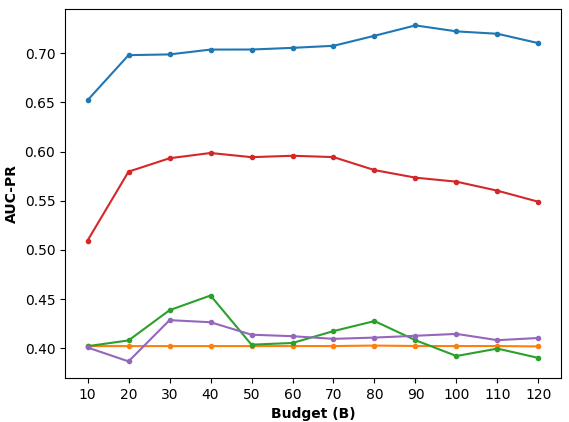}%
\caption{Synthetic Dataset}%
\label{subfiga}%
\end{subfigure}%
\begin{subfigure}{0.35\columnwidth}
\includegraphics[width=\columnwidth]{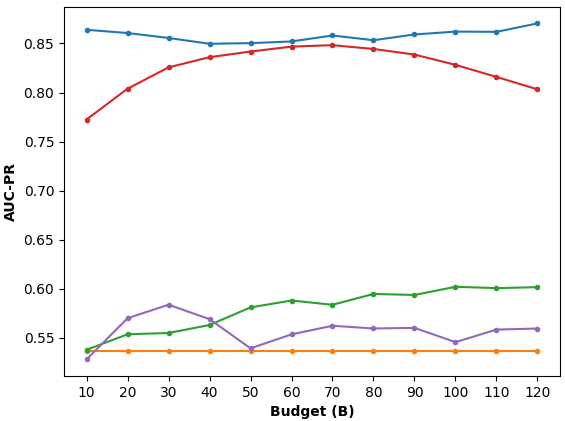}%
\caption{Synthetic Dataset}%
\label{subfiga}%
\end{subfigure}%

\begin{subfigure}{0.35\columnwidth}
\includegraphics[width=\columnwidth]{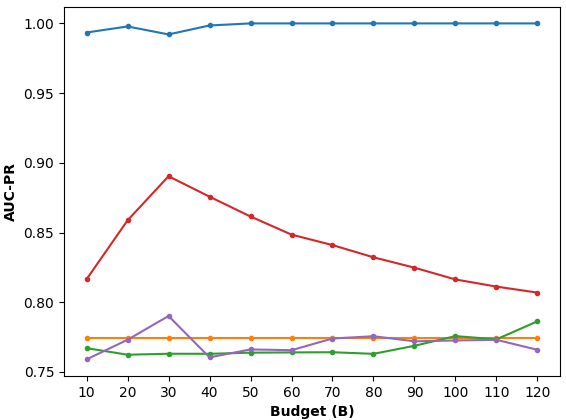}%
\caption{Synthetic Dataset}%
\label{subfiga}%
\end{subfigure}%
\begin{subfigure}{0.35\columnwidth}
\includegraphics[width=\columnwidth]{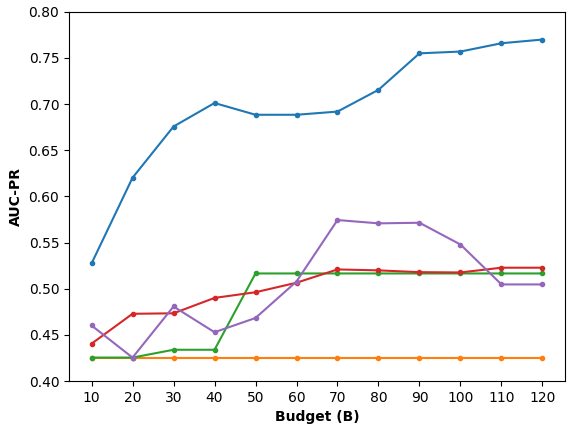}%
\caption{El Nino}%
\label{subfigb}%
\end{subfigure}%
\begin{subfigure}{0.35\columnwidth}
\includegraphics[width=\columnwidth]{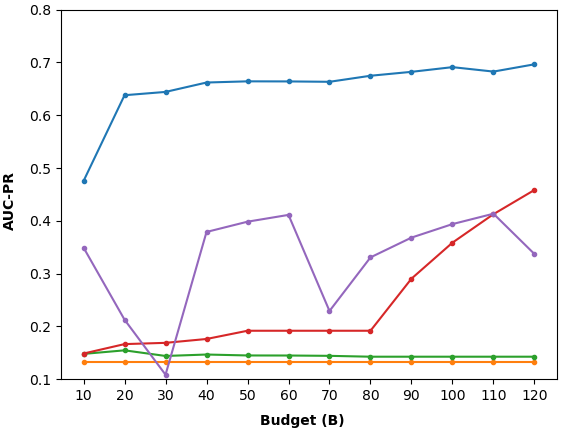}%
\caption{Houses}%
\label{subfigc}%
\end{subfigure}%

\begin{subfigure}{0.35\columnwidth}
\includegraphics[width=\columnwidth]{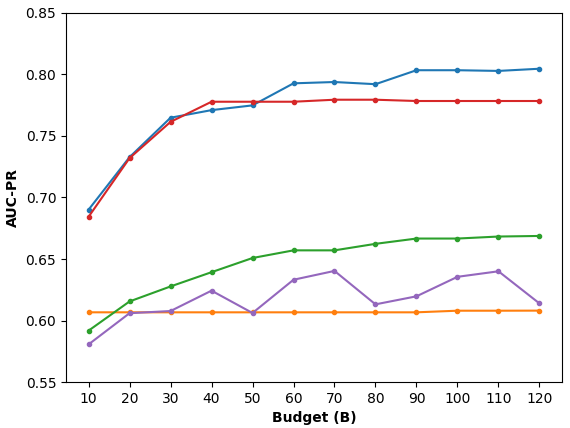}%
\caption{Abalone}%
\label{subfigg}%
\end{subfigure}%
\begin{subfigure}{0.35\columnwidth}
\includegraphics[width=\columnwidth]{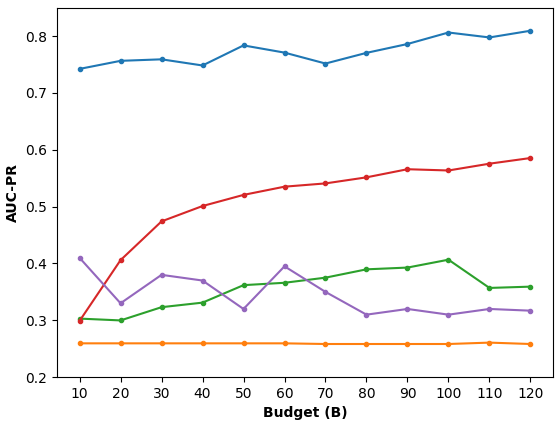}%
\caption{ANN-Thyroid}%
\label{subfigf}%
\end{subfigure}%
\begin{subfigure}{0.35\columnwidth}
\includegraphics[width=\columnwidth]{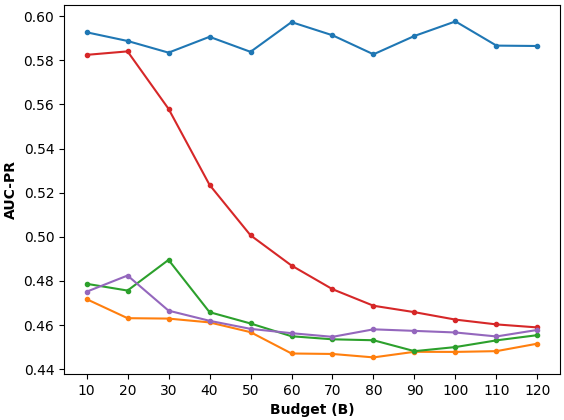}%
\caption{Arrhythmia}%
\label{subfig1}%
\end{subfigure}%

\begin{subfigure}{0.35\columnwidth}
\includegraphics[width=\columnwidth]{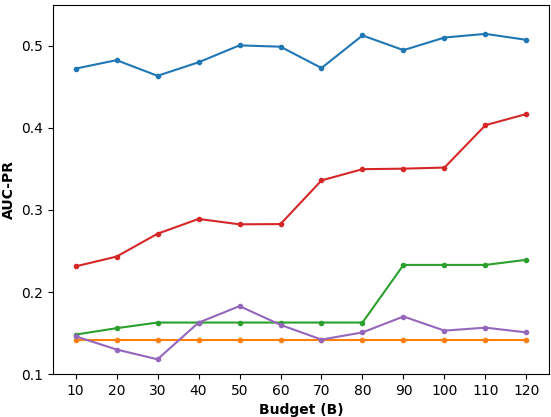}%
\caption{Mammography}%
\label{subfig4}%
\end{subfigure}%
\begin{subfigure}{0.35\columnwidth}
\includegraphics[width=\columnwidth]{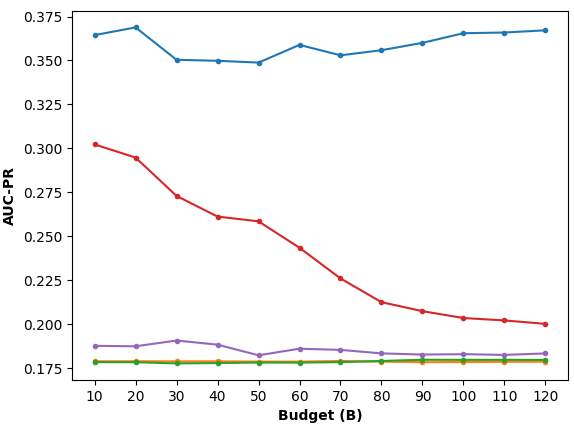}%
\caption{Letter}%
\label{subfig2}%
\end{subfigure}%
\begin{subfigure}{0.35\columnwidth}
\includegraphics[width=\columnwidth]{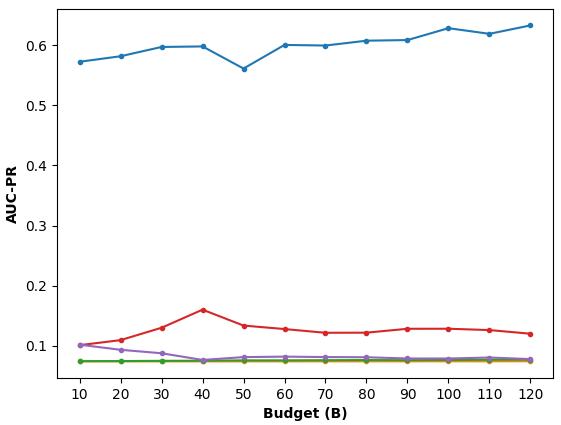}%
\caption{Optdigits}%
\label{subfig3}%
\end{subfigure}

\begin{subfigure}{0.35\columnwidth}
\includegraphics[width=\columnwidth]{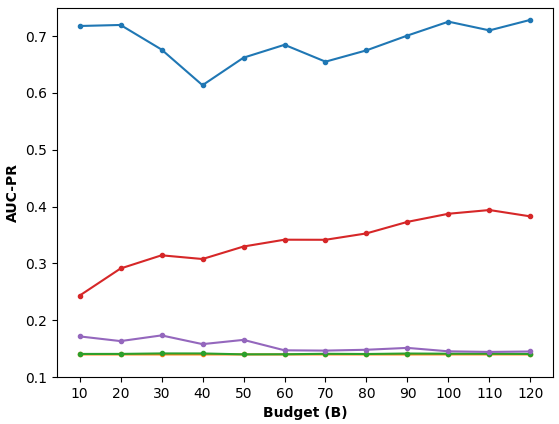}%
\caption{Pendigist}%
\label{subfigf}%
\end{subfigure}%
\begin{subfigure}{0.35\columnwidth}
\includegraphics[width=\columnwidth]{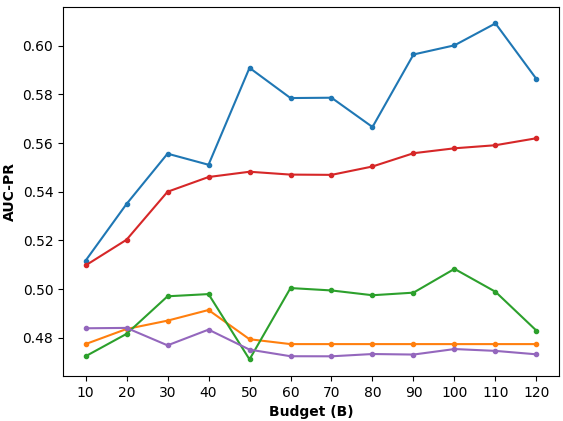}%
\caption{Satellite}%
\label{subfigf}%
\end{subfigure}%
\begin{subfigure}{0.35\columnwidth}
\includegraphics[width=\columnwidth]{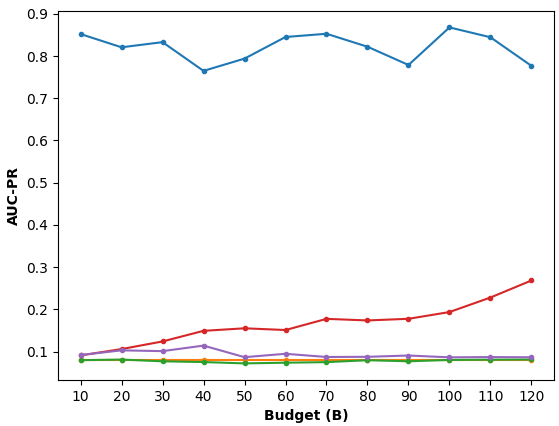}%
\caption{Satimage}%
\label{subfigf}%
\end{subfigure}%

\begin{subfigure}{0.35\columnwidth}
\includegraphics[width=\columnwidth]{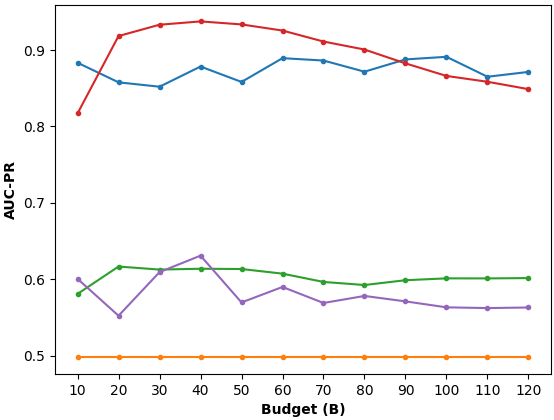}%
\caption{Thyroid}%
\label{subfigf}%
\end{subfigure}%
\begin{subfigure}{0.35\columnwidth}
\includegraphics[width=\columnwidth]{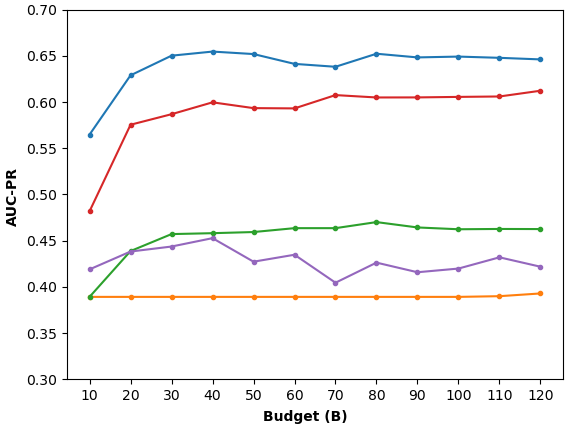}%
\caption{Vowels}%
\label{subfige}%
\end{subfigure}%
\begin{subfigure}{0.35\columnwidth}
\includegraphics[width=\columnwidth]{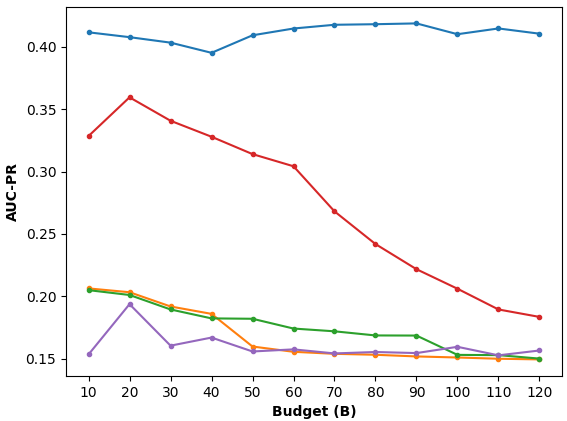}%
\caption{yeast}%
\label{subfigf}%
\end{subfigure}

\begin{subfigure}{0.2\columnwidth}
\includegraphics[width=\columnwidth]{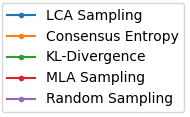}%
\end{subfigure}%
\caption{The performance (AUC-PR) vs. budget (b) for all datasets.}
\label{figbudgets}
\end{figure*}

\subsection{Results Q2: Comparison of the Query Strategies}\label{Q2}
In this experiment, we compare the performances of different query strategies described in Section \ref{queries}. 

Fig. \ref{figbudgets} shows the performances of the query strategies under different budgets. It can be seen that LCA sampling can quickly boost WisCon's performance, even using very few queries. Its performance curve flattens at a budget of less than 100 in most cases. It suggests that our query strategy is budget-friendly, meaning that it does not require too many iterations to find informative instances, even dealing with relatively large datasets such as El Nino.

On the other hand, the QBC strategies, i.e., consensus entropy and KL-divergence, perform poorly under reasonable budgets and do not provide major improvements over random sampling in most cases. Especially consensus entropy intensely suffers from the class imbalance and often fails to sample any instances from the positive class unless the budgets are really high. It is not uncommon for the learning with skewed distributions, where many otherwise useful active learning strategies may not be able to catch interesting minority instances \citep{settles2012active}. On the other hand, random sampling is the most unstable approach in terms of performance changes across different budgets.


Even though the MLA sampling, which explicitly targets anomalous instances, offers a significant improvement over these methods, it still consistently falls behind LCA sampling in majority of datasets. This shows that maximizing the number of true anomalies presented to the domain expert is not an appropriate objective. Unlike other works employing active learning with anomaly detection—e.g., \cite{das2016incorporating}, \cite{ lamba2019learning}, we cannot achieve an effective sampling with only this objective in our problem setting. This is mainly because the majority of the contexts are not beneficial in discovering the hard-to-find anomalies (see Section \ref{Q3}.), even if they do discover some of the easy ones. In the end, the anomalies that gain high confidence from the majority do not particularly help to reveal actual relevant contexts.

The experiments in this section prove that it is crucial to use an active learning strategy that suits the problem at hand. To effectively estimate the importance scores of different contexts, the strategy should be able to deal with not only the imbalance, but also with picking useful instances among rare anomalies. As a result, our proposed strategy, LCA sampling, significantly outperforms existing query models in all datasets.

\subsection{Results Q3: Justification of Ensembles and Active Learning} \label{Q3}
In this experiment, we verify the effectiveness, independently, of two main novelties in the WisCon: (1) context ensembles and (2) active learning. Our goal is to showcase how these two properties are complementary to each other in terms of providing better detection performance. 

First, we investigate the benefits of leveraging ensembles—that is, building block 3 in Fig. \ref{framework}—over using a single context. To this end, Table \ref{tab_ensemble_active} compares the performances of original WisCon, WisCon-Single, and WisCon-True. WisCon-Single is a variation of WisCon, which only uses the context assigned with the highest importance score instead of an ensemble. For both WisCon and WisCon-Single, LCA sampling with a budget of $100$ queries is used to estimate the importance scores. WisCon-True, on the other hand, represents the version of WisCon that uses ``true'' contexts for each dataset (see Section \ref{Datasets}). WisCon-Single represents the performance that can be achieved by a single context using active learning, while WisCon-True serves as the benchmark for the maximum performance that WisCon can achieve with a single context. 

The results show the clear benefit of using an ensemble of contexts over just selecting a single one. Even when the importance scores are estimated using an effective query strategy, WisCon-Single may significantly fall behind WisCon and WisCon-True in most of the cases. On the other hand, the original WisCon showcases ``the power of many'' by outperforming WisCon-Single, and even WisCon-True, in majority of the datasets. The clear difference can be especially observed in Synthetic 2 dataset comprising anomalies from multiple different contexts. We do not have the knowledge of characteristics of anomalies in real-world datasets, however, we can infer from these results that multiple useful contexts exist for the majority of them since WisCon can outperform WisCon-True in most of the cases. 

\begin{table*}[t]
\caption{Performance (AUC-PR) comparisons of WisCon with unsupervised ensembles and single-context baselines on seven datasets. The best performances are shown in bold.}
\begin{center}
\renewcommand{\arraystretch}{1.2}
\begin{tabular} {@{\extracolsep{0pt}}l|c|cc|cc}
\hline

\hline
&  \multicolumn{1}{c|}{} &\multicolumn{2}{c|}{\textbf{Single Context}}&\multicolumn{2}{c}{\textbf{Unsupervised Ensemble}} \\\cline{3-4} \cline{3-4} \cline{5-6}\cline{5-6}

\textbf{Dataset} & \textbf{\textit{WisCon}}  & \textbf{\textit{WisCon-Single}} & \textbf{\textit{Wiscon-True}}& \textbf{\textit{Averaging}} & \textbf{\textit{Maximization}}\\
\hline

\hline
Synthetic 1& $\mathbf{0.88 \pm 0.01}$ & $0.81 \pm 0.0$ & $0.81 \pm 0.0$ & $0.78 \pm 0.01$ & $0.39 \pm 0.0$\\
Synthetic 2& $\mathbf{0.72 \pm 0.06}$ & $0.27 \pm 0.14$ & $0.50 \pm 0.07$ & $0.40 \pm 0.08$ & $0.20 \pm 0.05$\\
Synthetic 3 &$\mathbf{0.86 \pm 0.04}$ & $0.30 \pm 0.14$ & $0.67 \pm 0.05$ & $0.53 \pm 0.04$ & $0.33 \pm 0.06$\\
Synthetic 4& $\mathbf{1.0 \pm 0.0}$ & $0.70 \pm 0.12$ & $0.96 \pm 0.06$ & $0.77 \pm 0.05$ & $0.99 \pm 0.01$\\
El Nino &$\mathbf{0.76 \pm 0.01}$& $0.71 \pm 0.02$& $0.62 \pm 0.05$ &$0.42 \pm 0.01$ &$0.35 \pm 0.03$ \\
Houses &$\mathbf{0.69 \pm 0.04}$ & $0.39 \pm 0.08$ & $0.63 \pm 0.06$ &$0.12 \pm 0.18$ & $0.08 \pm 0.21$ \\
Abalone &$\mathbf{0.81 \pm 0.09}$ &$0.62 \pm 0.24$ & $0.66 \pm 0.11$ &$0.56 \pm 0.13$ &$0.25 \pm 0.12$ \\
Ann-Thyroid &$\mathbf{0.80 \pm 0.04}$ & $\mathbf{0.80 \pm 0.04}$ & $\mathbf{0.80 \pm 0.04}$ &$0.32 \pm 0.01$ &$0.29 \pm 0.02$ \\
Arrhythmia &$\mathbf{0.60 \pm 0.09}$ &$\mathbf{0.60 \pm 0.09}$ & $0.54 \pm 0.08$ &$0.44 \pm 0.08$ &$0.48 \pm 0.12$ \\
Letter &$\mathbf{0.36 \pm 0.08}$ &$0.28 \pm 0.09$ & $0.33 \pm 0.08$ &$0.17 \pm 0.03$ &$0.23 \pm 0.04$ \\
Mammography &$\mathbf{0.50 \pm 0.05}$& $0.45 \pm 0.03$ & $0.45 \pm 0.03$ &$0.14 \pm 0.02$ & $0.09 \pm 0.0$\\
Optdigits &$\mathbf{0.62 \pm 0.08}$ &$0.15 \pm 0.12$ & $0.40 \pm 0.05$ &$0.07 \pm 0.01$ &$0.05 \pm 0.01$ \\
Pendigits &$\mathbf{0.72 \pm 0.09}$ &$0.32 \pm 0.07$ & $0.41 \pm 0.07$ &$0.14 \pm 0.03$ &$0.10 \pm 0.02$ \\
Satellite &$0.60 \pm 0.04$ &$0.59 \pm 0.01$ & $\mathbf{0.62 \pm 0.01}$ &$0.46 \pm 0.13$ &$0.50 \pm 0.01$ \\
Satimage &$0.86 \pm 0.08$ &$0.29 \pm 0.28$ & $\mathbf{0.90 \pm 0.05}$ &$0.07 \pm 0.03$ &$0.36 \pm 0.03$ \\
Thyroid &$\mathbf{0.89 \pm 0.05}$ &$0.74 \pm 0.04$ & $0.80 \pm 0.03$ &$0.50 \pm 0.08$ &$0.37 \pm 0.04$ \\
Vowels &$\mathbf{0.65 \pm 0.08}$ &$0.42 \pm 0.04$ & $0.48 \pm 0.06$ &$0.38 \pm 0.06$ &$0.17 \pm 0.03$ \\
Yeast &$\mathbf{0.41 \pm 0.10}$ &$0.30 \pm 0.09$ & $0.38 \pm 0.09$ &$0.14 \pm 0.03$ &$0.30 \pm 0.09$ \\
\hline

\hline
\end{tabular}
\label{tab_ensemble_active}
\end{center}
\end{table*}

Next, we show the effectiveness of using an active learning approach (i.e., building block 2 in Fig. \ref{framework}) in the ensemble combination process compared to classical unsupervised combination approaches. The second row in Table \ref{tab_ensemble_active} presents simple yet common combination strategies, i.e., averaging and maximization, that take the average and the maximum of the anomaly scores across all contexts. It can be seen that these two methods show significantly lower performances than WisCon, and are inferior even to WisCon-Single and WisCon-True context. This demonstrates that just incorporating multiple-contexts does not guarantee performance improvement over a single-context setting. It proves that context ensembles are only useful when they are done in an informed way, and active learning provides a good basis for this by effectively distinguishing useful contexts from irrelevant ones. 

Taking the maximum of anomaly scores could be considered a suitable ensemble combination method for WisCon as we claim that one dataset may contain multiple types of contextual anomalies and different contexts are complementary to each other. For example, the Con-Out \citep{meghanath2018conout}, which is the only prior work proposed as an ensemble over contexts, uses such a maximization technique to combine scores from multiple contexts. However, this strategy directly conveys errors (i.e., false positives) of each individual context to the ensemble's final results \citep{zimek2014ensembles}. Noting that WisCon assigns importance scores to the contexts based on their capability of unveiling actual (i.e., queried) anomalies, our pruning strategy may not eliminate the ones producing many false alarms. Furthermore, we do not claim that one type of contextual anomaly is only visible in a single context; however, different types may not stand out in the same context. Therefore, we chose weighted averaging for WisCon because it led to better performances and more stable results than the maximization of the scores in most datasets when we tested them both.

\begin{figure*}%
\centering
\begin{subfigure}{0.5\columnwidth}
\includegraphics[width=\columnwidth]{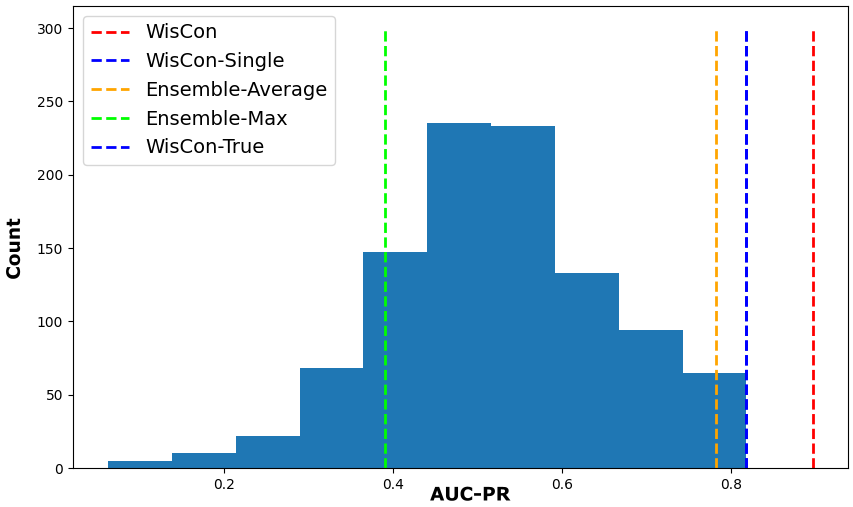}%
\caption{Synthetic 1 Dataset}%
\label{subfiga}%
\end{subfigure}%
\begin{subfigure}{0.5\columnwidth}
\includegraphics[width=\columnwidth]{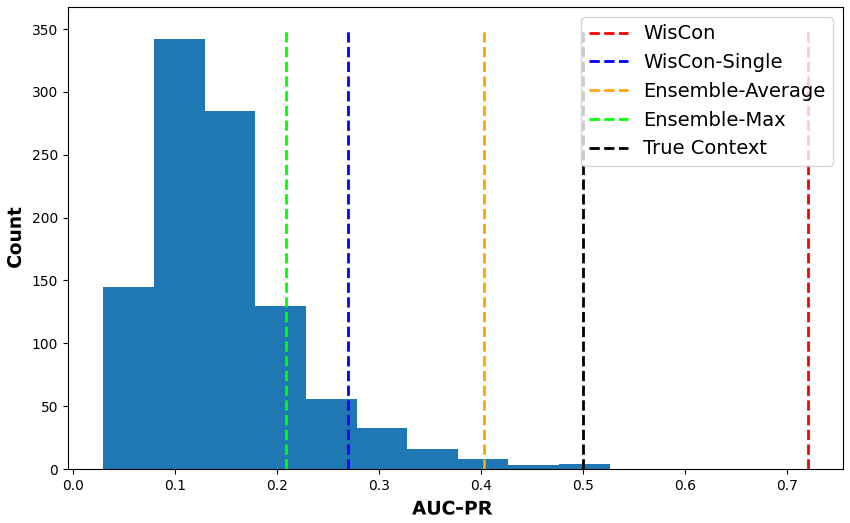}%
\caption{Synthetic 2 Dataset}%
\label{subfiga}%
\end{subfigure}%

\begin{subfigure}{0.5\columnwidth}
\includegraphics[width=\columnwidth]{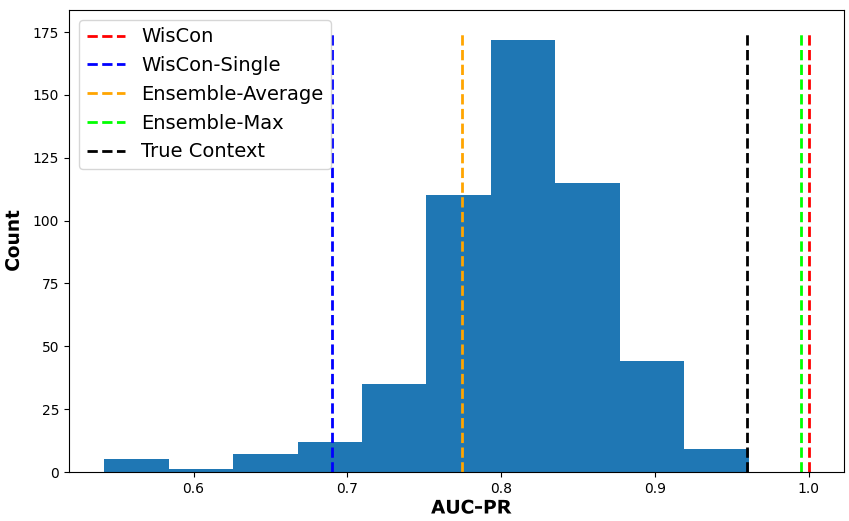}%
\caption{Synthetic 4 Dataset}%
\label{subfiga}%
\end{subfigure}%
\begin{subfigure}{0.5\columnwidth}
\includegraphics[width=\columnwidth]{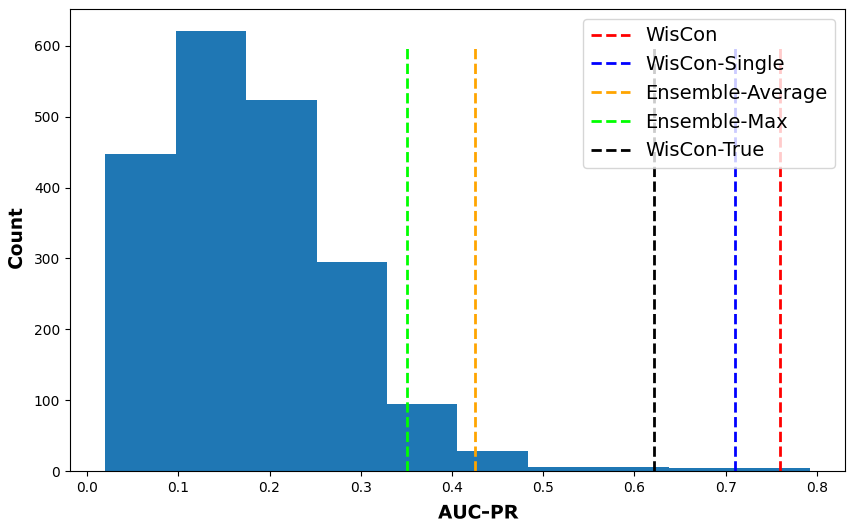}%
\caption{El Nino}%
\label{subfigb}%
\end{subfigure}%

\begin{subfigure}{0.5\columnwidth}
\includegraphics[width=\columnwidth]{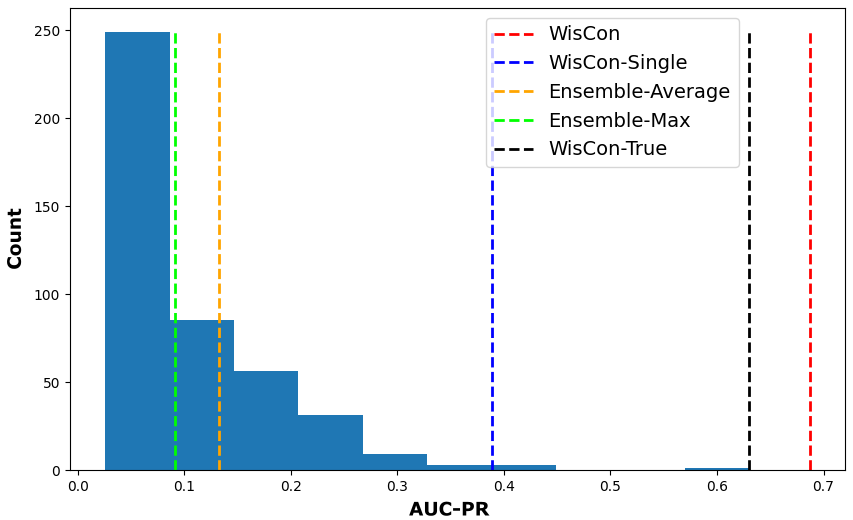}%
\caption{Houses}%
\label{subfigc}%
\end{subfigure}%
\begin{subfigure}{0.5\columnwidth}
\includegraphics[width=\columnwidth]{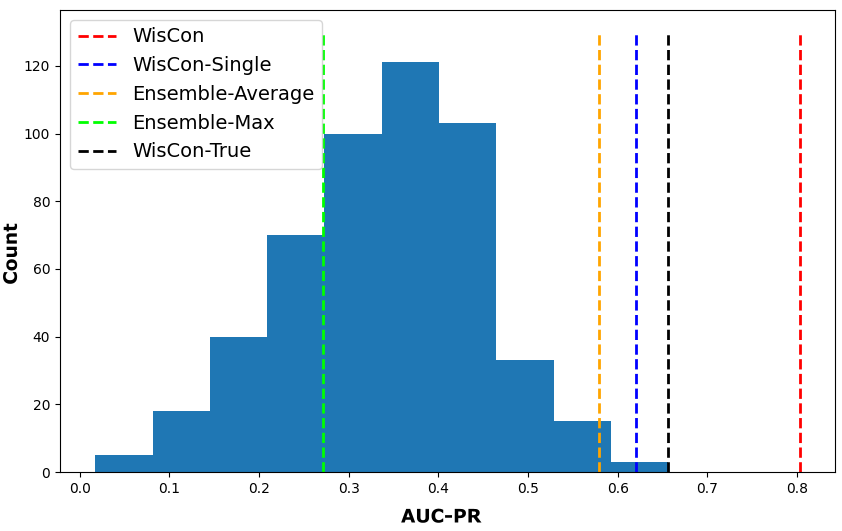}%
\caption{Abalone}%
\label{subfigg}%
\end{subfigure}%

\begin{subfigure}{0.5\columnwidth}
\includegraphics[width=\columnwidth]{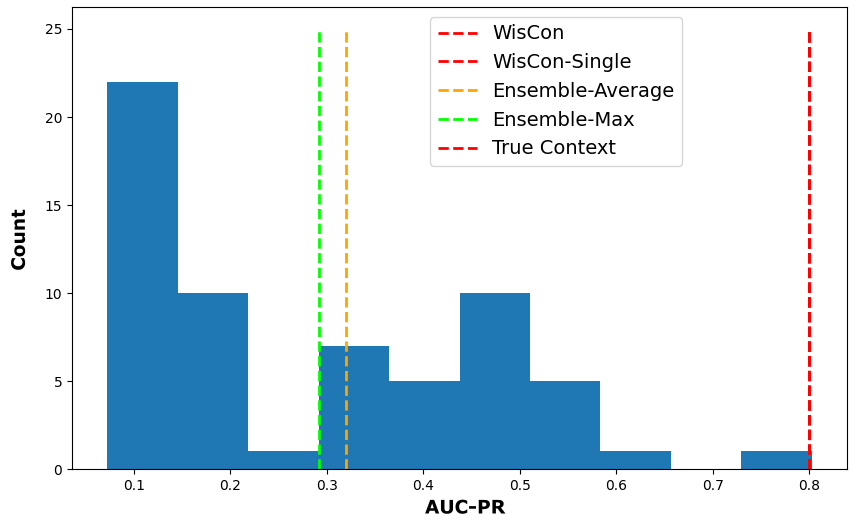}%
\caption{ANN-Thyroid}%
\label{subfigf}%
\end{subfigure}%
\begin{subfigure}{0.5\columnwidth}
\includegraphics[width=\columnwidth]{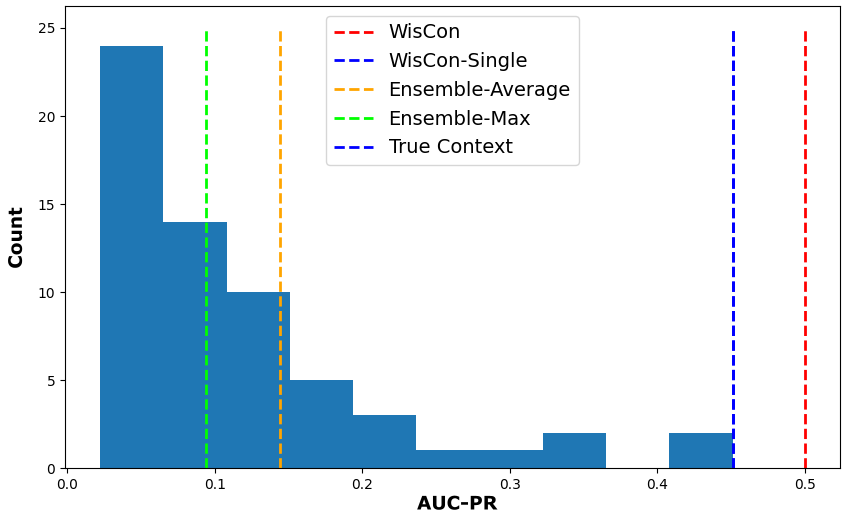}%
\caption{Mammography}%
\label{subfigd}%
\end{subfigure}\hfill%

\begin{subfigure}{0.5\columnwidth}
\includegraphics[width=\columnwidth]{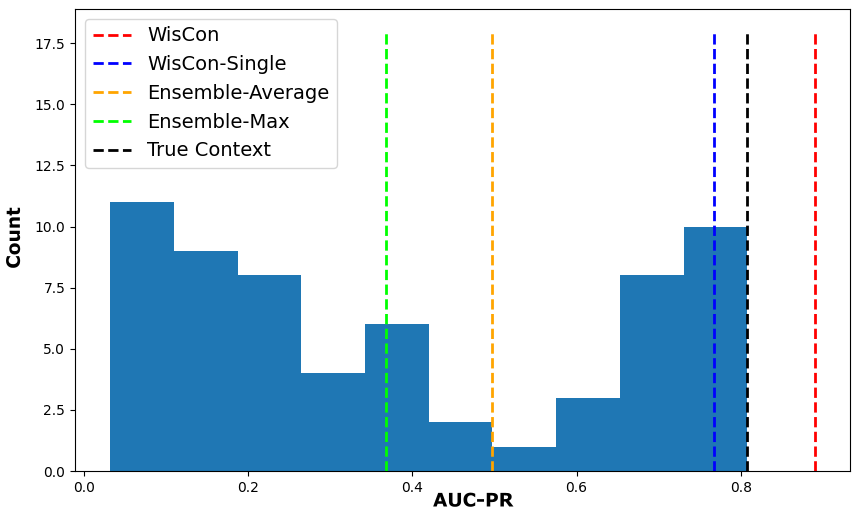}%
\caption{Thyroid}%
\label{subfigf}%
\end{subfigure}%
\begin{subfigure}{0.5\columnwidth}
\includegraphics[width=\columnwidth]{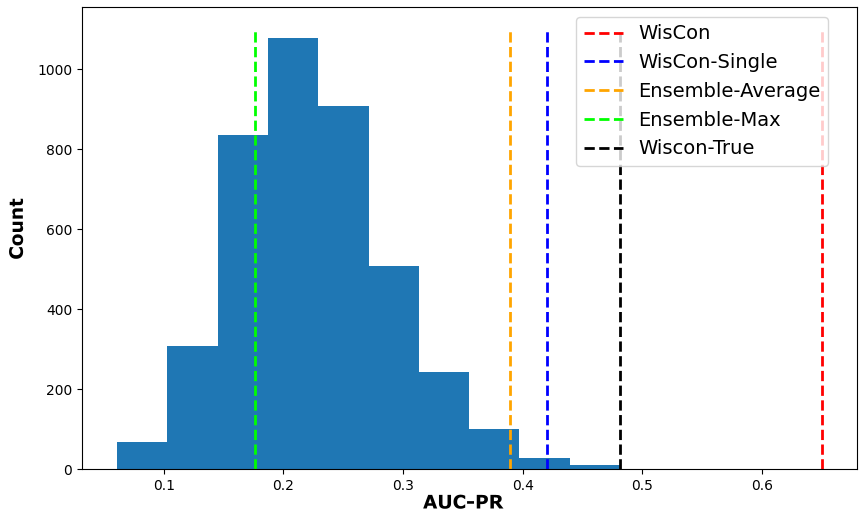}%
\caption{Vowels}%
\label{subfige}%
\end{subfigure}%

\begin{subfigure}{0.5\columnwidth}
\includegraphics[width=\columnwidth]{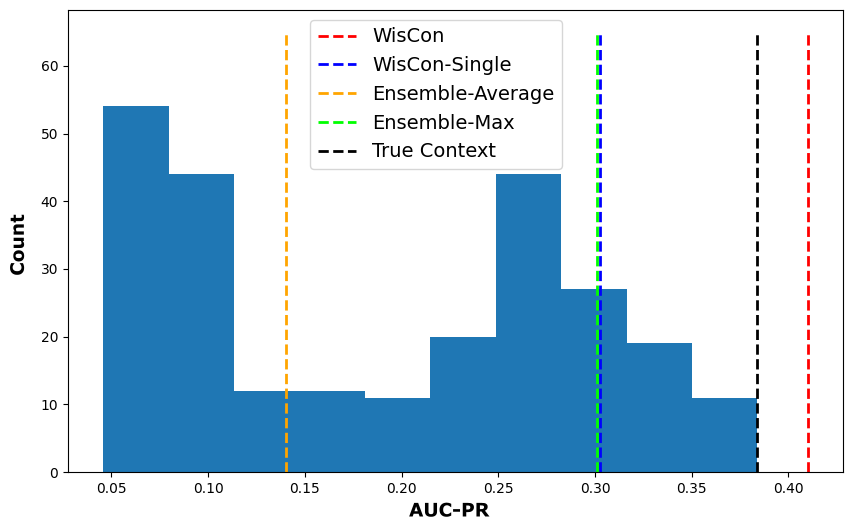}%
\caption{Yeast}%
\label{subfige}%
\end{subfigure}%

\caption{The distribution of the performances (AUC-PR) in different contexts for all datasets.}
\label{figabc}
\end{figure*}

Fig. \ref{figabc} further shows the distribution of AUC-PR scores across all contexts in each dataset, along with the performances of different approaches reported in Table \ref{tab1}. The majority of the distributions are skewed towards the left, showing that most contexts in the entire population lead to poor performances. This observation supports our initial hypothesis when designing our query strategy, the LCA sampling, in Section \ref{lca}. It also shows the importance of including a pruning strategy in our final ensemble since the poor contexts may deteriorate the overall performance by ``overwhelming'' the good contexts.

These figures also demonstrate the complexity of specifying the true context in practice. It can be seen that the number of possible contexts increases dramatically when the number of attributes is high. Therefore, it is not feasible for a human expert to carefully consider many different combinations of all attributes to determine the best possible context. In addition, unexpected behavior can be observed regarding El Nino dataset. The contextual anomalies injected into this dataset are the instances perturbed under a pre-defined context (See section \ref{Datasets}). Technically, the true context that reveals the contextual anomalies best should be the context used during the perturbation. However, Fig. \ref{subfigb} shows that there are some outliers on the right side of the distribution that resulted in higher performances than the actual true context. This phenomenon showcases the risk of relying on a single context even when we obtain a certain level of knowledge on what the roles of different attributes are supposed to be in that system.      

Regardless of the benefits of WisCon that we have demonstrated throughout this section with different experiments, the proposed approach certainly has a number of weaknesses to be aware of. First, WisCon's active learning mechanism displays a clear drawback against unsupervised baselines which do not require \textit{any} labels. Even though WisCon provides superior detection performance in most datasets with very few labels, the requirement for labels limits its applicability. Another important drawback is handling high-dimensional datasets. As we claim in the introduction and show throughout Section 6, useful contexts are rare, and the number of possible contexts grows exponentially with the number of features. It is not possible for WisCon to directly scale to high-dimensional datasets without pre-processing. Nevertheless, other approaches to automatically lower the number of contexts should be considered in the future for the cases where PCA does not work sufficiently well. Finally, WisCon uses a straightforward yet inefficient base detector that creates separate iForest per reference group. It increases computational effort and makes estimating optimum hyper-parameters for multiple models simultaneously difficult. In this work, we achieved reasonable performances, despite only running the base detector with the default parameters recommended for iForest \citep{liu2008isolation}. 


\section{Conclusion}\label{Conclusion}
In this work, we addressed the problem of contextual anomaly detection where ``true'' contextual and behavioral attributes are unknown. We have introduced the Wisdom of the Contexts (WisCon) approach, bringing together the best of two worlds, namely active learning and ensembles, to deal with complex contextual anomalies. Motivated by the assumption that not all contexts created from the feature space are equally advantageous in identifying different contextual anomalies, WisCon accommodates a measure to quantify their usefulness (i.e., importance) using an active learning schema. Instead of relying on a pre-specified single one, our approach maintains an ensemble by combining multiple contexts according to their importance scores.   

We have also proposed a new committee-based query strategy, low confidence anomaly (LCA) sampling, designed to select the best samples to accurately distinguish between ``relevant'' and ``irrelevant'' contexts so that the importance scores can be estimated within a small budget.

We have conducted several experimental studies to demonstrate the benefits of our WisCon approach and proposed query strategy. First, we have showcased the efficacy of the approach against 16 different baselines in three different categories (i.e., active learning baselines, unsupervised anomaly detectors, unsupervised contextual anomaly detectors) with two different performance metrics. WisCon outperformed its competitors from all the categories in majority among 18 datasets that contain both artificial and real-world anomalies.  

We have then compared different query strategies under varying budgets and demonstrated that commonly used methods are not suitable for this problem. LCA sampling not only consistently outperforms other approaches but also provides higher cost-efficiency by requiring lower budgets.

Finally, we have shown the individual impact of the two key novelties of WisCon, the active learning and context ensembles. Our experiments have proven the initial hypothesis that there is no single perfect context that successfully uncovers all contextual anomalies, and also have showcased the clear advantage of ``the wisdom of the contexts'' over an ``individual'' one. However, it has also revealed that just incorporating multiple contexts does not ensure superiority over a single context. Our WisCon approach can significantly boost detection performance by effectively building ensembles using active learning with a proper query strategy.

\bibliographystyle{spbasic}      
\bibliography{template}

\end{document}